\documentclass[10pt,twocolumn,letterpaper]{article}

\usepackage{cvpr}              

\usepackage{graphicx}
\usepackage{array}
\usepackage{amsmath}
\usepackage{amssymb}
\usepackage{booktabs}
\usepackage{longtable}
\usepackage{tabularx}
    \newcolumntype{L}[1]{>{\raggedright\arraybackslash}p{#1}}

\usepackage{algorithm}
\usepackage{algpseudocode}
\usepackage{listings}
\usepackage{xcolor}
\lstdefinestyle{prompt}{
  basicstyle=\ttfamily\small,
  backgroundcolor=\color{gray!10},
  frame=single,
  breaklines=true,
  postbreak=\mbox{\textcolor{red}{$\hookrightarrow$}\space},
}

\usepackage{float}
\usepackage{array, makecell}
\usepackage{multirow, multicol}
\usepackage{enumitem}
    \setlist[itemize]{leftmargin=*, labelindent=2em}

\definecolor{cvprblue}{rgb}{0.21,0.49,0.74}
\usepackage[pagebackref,breaklinks,colorlinks,allcolors=cvprblue]{hyperref}

\usepackage{longtable}
\newcounter{supfig}

\newcounter{suptab}

\def\paperID{13} 
\def\confName{CVPR-CogVL}
\def\confYear{2026}

\title{Action Without Interaction: Probing the Physical Foundations of Video LMMs via Contact-Release Detection}

\author{
    Daniel Harari$^1$ \quad Michael Sidorov$^1$ \quad Chen Shterental$^1$ \quad Liel David$^1$ \\
    Abrham Kahsay Gebreselasie$^2$ \quad Muhammad Haris Khan$^2$\\[6pt]
    $^{1}$Weizmann Institute of Science\\
    $^{2}$Mohamed bin Zayed University of Artificial Intelligence\\[4pt]
    {\tt\small \{hararid, michael.sidorov, chen.shterental\}@weizmann.ac.il, olesya.liel@gmail.com}\\ 
    {\tt\small \{abrham.gebreselasie, muhammad.haris\}@mbzuai.ac.ae}\\
}

\begin{document}
\maketitle
\begin{abstract}
Large multi-modal models (LMMs) show increasing performance in realistic visual tasks for images and, more recently, for videos. For example, given a video sequence, such models are able to describe in detail objects, the surroundings and dynamic actions. In this study, we explored the extent to which these models ground their semantic understanding in the actual visual input. Specifically, given sequences of hands interacting with objects, we asked models when and where the interaction begins or ends. For this purpose, we introduce a first of its kind, large-scale dataset with more than 20K annotated interactions on videos from the Something-Something-V2 dataset. 250 AMTurk human annotators labeled core interaction events, particularly when and where objects and agents become attached (`contact') or detached (`release'). We asked SoTA LMMs, including GPT, Gemini and  Qwen to locate these events in short videos, each with a single event. The results show that while models reliably name target objects and identify actions, they exhibit a form of `shortcut learning' where semantic success masks a failure in physical grounding.
Specifically, they consistently fail to identify the frame where the interaction begins or ends and poorly localize the physical event within the scene. This disconnect suggests that while LMMs excel at System 1 intuitive pattern recognition (naming the action and objects), they lack the System 2 cognitive foundations required to reason about physical primitives like `contact' and `release', hence truly ground dynamic scenes in physical reality.

\end{abstract}    

\section{Introduction}
\label{sec:intro}
\begin{figure}[t]
    \centering
        \includegraphics[width=1\columnwidth]{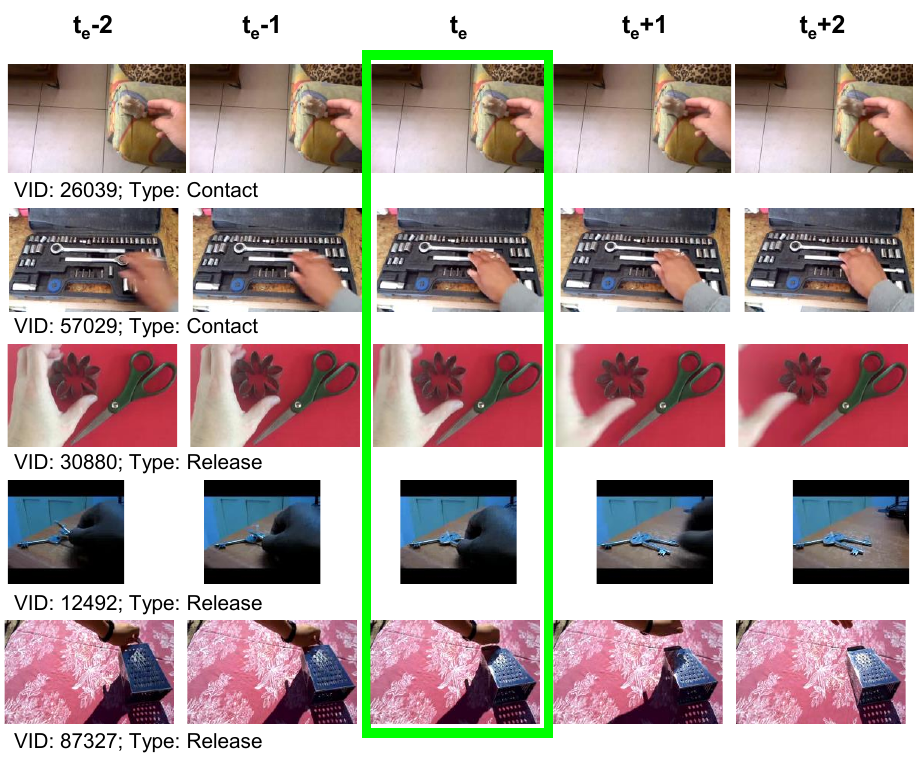}
    \caption{\textbf{Events of physical contact or release.}  Each row presents 2 frames before and after a true physical contact or release event (marked with a green box) during a hand-object interaction. Humans are sensitive to the gap between the hand and the target object even one frame before a contact or after a release. Evidence of a physical contact is crucial to identify an interaction over a simple occlusion, or to predict the outcome of an interaction at it's termination. These examples are from our annotated dataset, extracted from SSv2.}
    \label{fig:example_events}
\end{figure}

Discovering and understanding actions and interactions are fundamental cognitive capabilities of humans and other intelligent beings, necessary in interpreting and planning dynamic events between objects and agents in the surrounding environment \cite{radvansky2014event}. Infants develop early sensitivity to spatiotemporal continuity in simple events, for example, the physical contact between an agent and a target object (see \cref{fig:example_events}). 
In developmental psychology, the perception of physical contact and release are considered `primitives' of causal reasoning—foundational building blocks that allow humans to move beyond surface correlations to an understanding of agency and physical laws.
This sensitivity to the causality of perception guides infants at a very young age, in learning to detect and interpret interactions between objects and agents, including launching, entraining and expulsion events \cite{michotte1963perception, leslie1982causality, saxe2006perception, baillargeon1995physical}. 

Computationally, recent vision models showed increasing performance in recognizing actions and interactions in realistic video sequences \cite{morshed_human_action_survey2023, gong_action_survey2025, tang_llm_video_survey2025}. Some models include special architectural designs to improve on the internal representation learning~\cite{long_videoprism2024}, while others use a common architecture, but train on very large unlabeled video datasets utilizing self-supervised learning (SSL) paradigms \cite{tong2022videomae, liu_vlm2023, ishan_videossl2024}. The introduction of large multi-modal models (LMMs), allowed the combination of semantic information from large language models (LLMs) and foundational visual representations, thus allowing to generalize to unseen videos and actions without explicit training~\cite{tang_llm_video_survey2025}.

\begin{figure*}[t]
    \centering
    \includegraphics[width=1.0\textwidth]{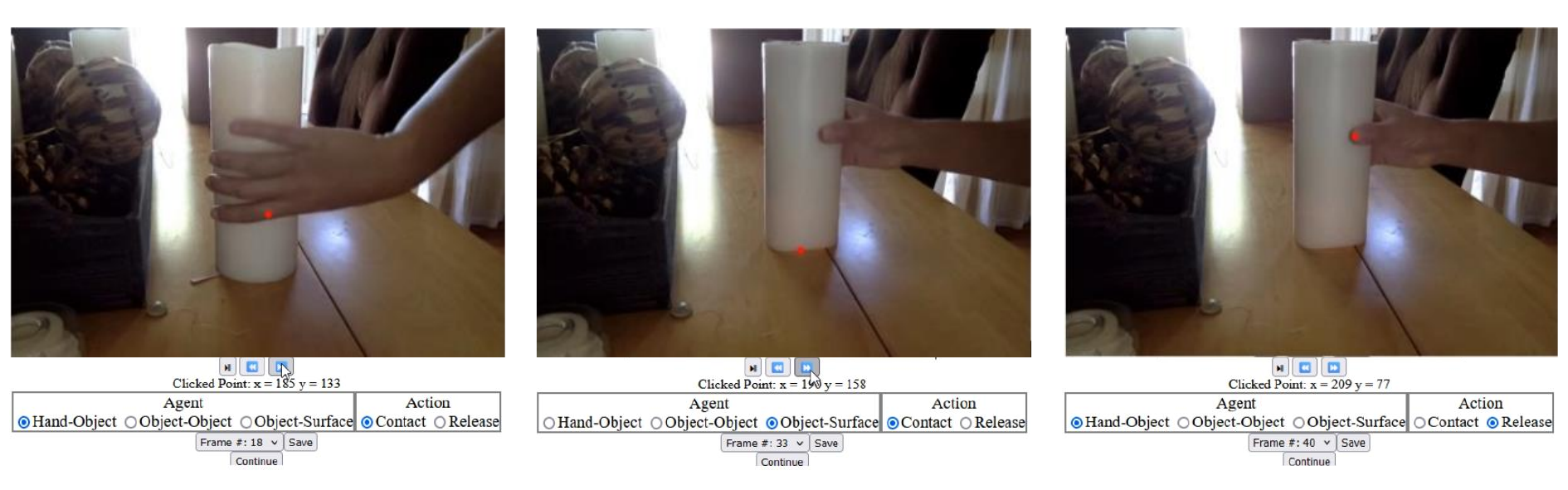}
    \caption{\textbf{Collecting human annotations for interactions using Amazon Mechanical Turk platform.} Human subjects were asked to annotate core interaction events in videos from SSv2 dataset~\cite{goyal2017something}. Shown here are example annotations for `contact' and `release' events, where the target object (white candle) comes in contact with a hand (left) and a surface (middle), or is detached from the hand (right). The annotations include the event type, the kind of agent-object pair and the spatiotemporal location of the event (frame and image coordinates).}
    \label{fig:dataset_annot_survey}
\end{figure*}

Despite the increasing success of models in generalizing to more complex tasks without explicit training, recent studies revealed fundamental limitations in the models' ability to reason about the performed tasks and develop human-like generalizable cognitive understanding \cite{cherian_smart2023, buschoff_nature2025, hou2025do}. In this study, we explore whether the enhanced performance of LMMs in video action recognition, reflects an improved cognitive understanding (System 2), or merely a superficial `story telling' ability about detected objects in proximity of hands (System 1)?
    
For this purpose we introduce a large-scale dataset -- \textit{The Contact-Release Interaction Dataset} (CRID). This first of its kind dataset consists of more than 20K annotated interactions, based on $10,000$ action videos from the ``Something-Something v.2'' (SSv2) dataset~\cite{goyal2017something}. Using the Amazon Mechanical Turk (AMTurk) crowd sourcing platform, we conducted a survey, in which 250 human annotators labeled $24,222$ core interaction events, including: (i) the type of agent acting upon the target object (e.g., a hand or another object), (ii) the type of core interaction event (`contact'--attachment between a target object and an agent, or `release'--detachment of the object from the agent), (iii) the spatiotemporal location of the event (frame number and image coordinates). \cref{fig:dataset_annot_survey} depicts the annotation setting (see details in~\cref{sec:dataset}). 

Based on these annotations, we conducted a series of experiments to evaluate the ability of current LMMs to detect the spatiotemporal location of core interaction events in real-world video sequences. The experiments were conducted under several In-Context-Learning (ICL) regimes, while applying two modifying conditions on the models' prompts -- \textit{Reasoning} and \textit{Grounding} -- inspired by earlier studies \cite{shao_visual_cot2024, wei_cot_prompting2022, guo_deepseekr12025}. We applied the evaluation scheme (see \cref{fig:exp_flow_chart}) to five SoTA LMMs including, OpenAI's GPT-5.2 and GPT-4o models \cite{singh2025openai, openai2024gpt4o}, Google's Gemini-Pro-3 and Gemini-2.5-Flash models \cite{google_gemini_3_pro_2026, google_gemini_2_5_flash_2025} and an open source version of Alibaba's Qwen-2.5VL-72B \cite{Qwen2.5-VL}.

In summary our contributions include:
\begin{itemize}
    \item Introduce CRID - a large scale dataset, based on 10K interaction videos from SSv2 dataset, with more than 20K first of their kind human annotations of core interaction events (`Contact' and `Release'). The annotations include details on the event and agent types, as well as the spatiotemporal locations of the events.
    \item A set of prompting experiments under several ICL regimes and modifying \textit{Reasoning} and \textit{Grounding} conditions.
    \item A discussion around the striking `grounding gap' revealed in our results: models can identify a high-level action label through statistical association, but remain `blind' to the underlying physical events that define that action.
\end{itemize}

\section{Related Work}
\label{sec:related-work}

Video understanding was thoroughly researched in the past due to its high value to the advancement in the domain of AI. Unlike action recognition in video, such as identifying people jumping, playing tennis etc., video understanding is a more complex task, often requiring a high level of generalization. Recent studies in this area of research, such as Maaz et al. \cite{maaz2024video}, showed that a ChatGPT agent can successfully answer complex questions when prompted with image data together with the verbal question. 
Wu et al. \cite{wu2025f} used segmentation masks, which the model provided in a grounding step, to further improve the model's understanding of the input images.
Shao et al. \cite{shao_visual_cot2024} showed that by asking the model to produce a Chain-of-Thought of a general task related to the image of the main task, improves significantly the final answer of the LMM.
Tian and Wu \cite{tian2025llm} also used a prompt tuning technique to improve action recognition performance of an LMM agent, but their method requires training of additional adapter models that embed the prompt, actions and the image to be tokenized and sent to the LMM. 
Chen et al. \cite{chen2025visrl} utilize reinforcement learning to improve LMM grounding via iterative ROI extraction and prediction-based feedback. 

While Qi et al. \cite{Shuhan_llm_limit2023} reveal that image-based LMMs often prioritize textual priors over visual evidence, their analysis remains limited to static scenes. Our work extends this critique to the temporal domain, demonstrating that even when models produce coherent action narratives, they fail to ground them in the physical primitives—such as contact and release—that define dynamic interactions.

Finally, several studies focused on changes that a manipulated object undergoes during long interaction videos (e.g., paper folding, fruit peeling, etc.), utilizing existing benchmarks for object affordance and object manipulation~\cite{soucek2022lookforthechange, xue2023learning, mahiro2025bench, zameni2025moscato}. The videos are often at low resolution and include coarse categories of background and actions, rather than the begin or end of a physical contact.
In contrast, our evaluation is focused on the initial moment of contact or release in an interaction, regardless of the kind of target object or the kind of manipulation, which are the core of the object affordance studies. We believe that the generic ability to detect physical contact is fundamental for learning about object affordance and manipulations in a self-supervised manner.

\section{Dataset}
\label{sec:dataset}

\begin{figure*}[t]
    \centering
    \includegraphics[width=0.95\linewidth]{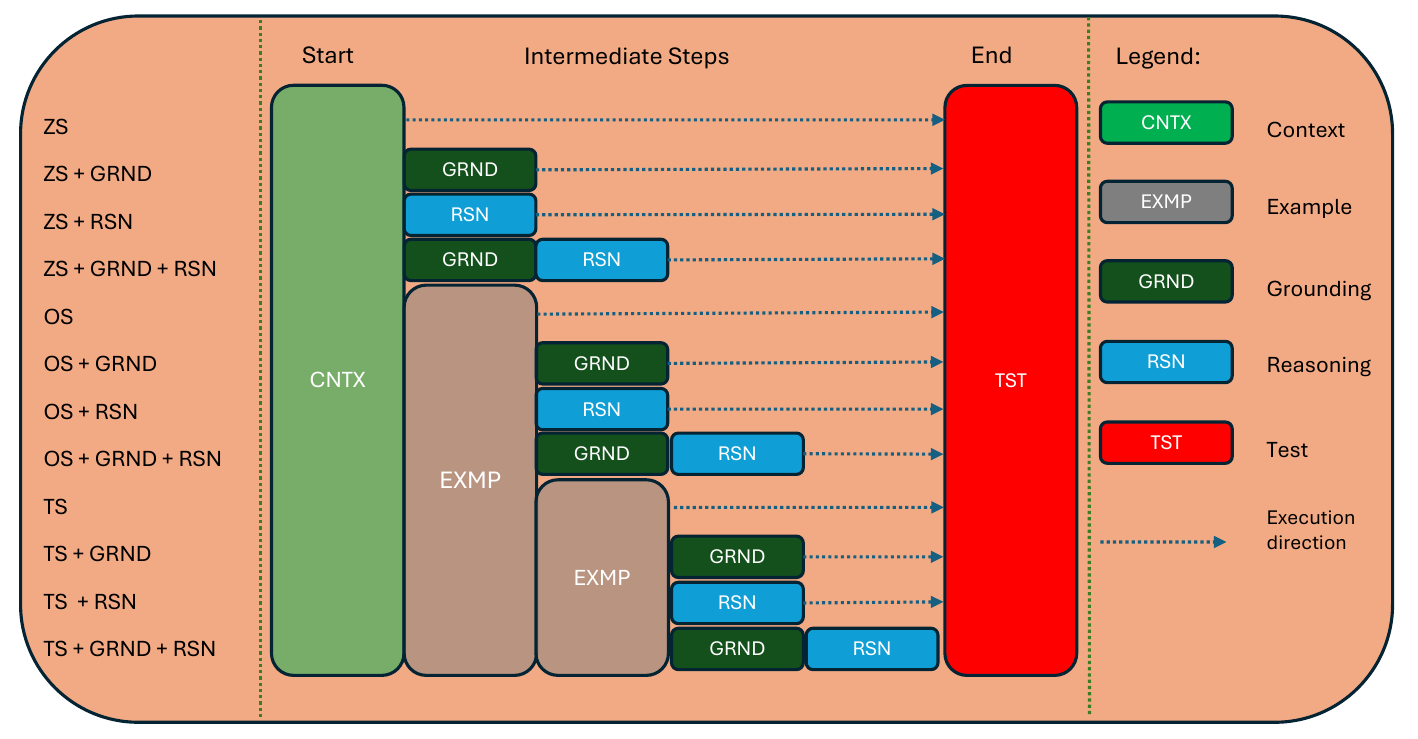}
    \caption{A schematic flow chart of the experiments under the different In-Context-Learning (ICL) regimes (i.e., ZS, OS, TS) and modulating conditions. The blocks represent different components of intermediate procedures. Each row represents an experiment using a particular ICL regime and condition (the experiment flow is directed left to right). The CNTX block indicates an introductory prompt about the agent. The EXMP block represents a prompt of an example, including the task instruction, an input video and the correct response for this example. The RSN block indicates a prompt instructing the model to include in the response a step-by-step description of the reasoning behind the predicted answer. The GRND block represents a prompt instructing the model to describe the content of the input video and the instructing prompt. In this block, the model provides an intermediate response, prior to the main task. The TST block indicates the prompt of the main test task, including the instruction and test video (see \cref{sec:experiments} for more details).}
    \label{fig:exp_flow_chart}
\end{figure*}
\paragraph{Overview.}
In this section we describe the dataset used in our experiments. 
We explore models' understanding of core interaction elements, for example the moment a hand picks up a target object and the spatial location where the physical contact occurs in the scene. We introduce a new large scale dataset -- \textit{The Contact-Release Interaction Dataset (CRID)}. This first of its kind dataset consists of more than 20K annotated interactions on more than 10K action videos from the ``Something-Something-v2'' (SSv2) dataset, which is a collection of smartphones' videos depicting humans performing various everyday interactions in natural settings~\cite{goyal2017something}. The SSv2 labels include generic action templates, for example, ``putting something into something'', and the object names in the template placeholders (``something'') for each video.

\setlength{\tabcolsep}{20pt}
\begin{table}[b]
    \caption{\textbf{\textit{The Contact-Release Interaction Dataset (CRID)}}. New annotations of core interaction events for videos extracted from the SSv2 dataset~\cite{goyal2017something}.}
    \centering
            \begin{tabularx}{1.0\linewidth}{lr}
            \toprule
                \textbf{Videos}                          & $10130$ \\
                \textbf{SSv2 action templates}          & $91$ \\
                \textbf{Mean videos per template}        & $111$ \\
                \midrule
                \textbf{`Contact' events}                & $13816$ \\
                \textbf{`Release' events}                & $10406$ \\
                \textbf{`Hand-object' interactions}      & $12550$ \\
                \textbf{`Object-object' interactions}    & $5653$ \\
                \textbf{`Object-surface' interactions}   & $6019$ \\
                \bottomrule
            \end{tabularx}
    \label{tab:dataset_summary}
\end{table}

\vspace{-1em}
\paragraph{Human annotations.} In this work we used the AMTurk online platform to conduct a survey, in which 250 human annotators labeled more than 20K core interaction events, including: (i) the type of agent acting upon the target object (e.g., a hand or another object), (ii) the type of core interaction event, i.e, `contact' or `release', (iii) the spatiotemporal location of the event, i.e., the frame number and image coordinates where the physical event occurs (see \cref{fig:example_events} and \cref{fig:dataset_annot_survey}). The survey and the annotations collection procedure were approved by the institutional review board of the Weizmann Institute of Science, Rehovot, Israel. All human subjects gave informed consent before participating in the survey.
Details about the new annotated dataset, including the videos and the labels are summarized in~\cref{tab:dataset_summary}.
Due to time and budget constraints, we could not hire multiple annotators for each of the 10K videos. However, we evaluated the inter-rater agreement between randomly selected annotators, by measuring the Intraclass Correlation Coefficient (ICC) between 3 random annotators on a randomly selected subset of videos. The results indicate a high agreement level on the frame annotation (ICC of 0.95), and a medium agreement level on the spatial location where the events occur (ICC of 0.73 and 0.39 for the exact x and y coordinates, respectfully, and an IoU of 0.57 for a 120$\times$120 bounding box centered around the labeled location).

\vspace{-1em}
\paragraph{Open access availability.} The dataset is publicly available at \href{https://gitfront.io/r/hararid/52Roq1ASsNte/ssv2-contact-release-interaction-dataset}{ssv2-contact-release-interaction-dataset}. Despite concerns about benchmark data leaking into LMM pre-training, recent research (e.g.,~\cite{Gunawardhana2024SelfSupervised}) indicates that even top-tier SSL visual encoders struggle with physical interactions and contact, even after fine-tuning. Consequently, we provide open access to our annotations. This allows future models to leverage these scarce examples alongside their inherent reasoning and semantic knowledge (similar to humans) to bridge current limitations.

\section{Experimental Design}
\label{sec:experiments}

\paragraph{Overview.} In this section we describe in more detail the evaluation experiments conducted in the course of this study. We considered three ICL regimes: zero-shot, one-shot and two-shot~\cite{dong_icl_survey2024}. 

As we are interested in models' ability to understand core interaction events in videos, we focused on the detection of the frame in the video, where a core interaction event (`contact' or `release') occurs. Since the original SSv2 videos often contain several simultaneous interaction events, we manually extracted 3 temporally isolated events from a subset of 33 videos, resulting with 99 short sequences, 10 frames each, consisting of a single core event (see supplementary for details). As a control, we extracted from each video, an additional short sequence (a `non-event'), which depicts objects and hands in a dynamic scene without a physical contact/release event.

To better understand the models behavior, we employed two explainability methods, that were used in previous studies with LMMs, as modulating conditions: (i) \textit{Grounding}, (ii) \textit{Reasoning}. We elaborate about these conditions later in this section. In our experiments we evaluated their relative influence on the models performance. These prompting conditions do not modify the models' internal reasoning parameters. \cref{fig:exp_flow_chart} presents the different experimental settings, including the ICL regimes and modulating conditions.

\vspace{-1em}
\paragraph{Zero-Shot (ZS) regime.} In the baseline ZS regime, models are instructed to perform the main task on a test video (i.e., detect the frame where the a core event occurs), without any examples. The prompt includes an introduction, the image frames of the test video sequence and the instructions for the main task, as shown in \cref{lst:baseline_prompt}.

\begin{figure}[t]
\begin{lstlisting}[style=prompt, basicstyle=\scriptsize\ttfamily, escapeinside={(*@}{@*)}, caption={The prompt used in the baseline ZS experiment.}, label={lst:baseline_prompt}]
(*@\textbf{System:}@*) 
You are a useful assistant and an expert in video understanding.
(*@\textbf{Images:}@*) 
[First test video frame] 
...
[Last test video frame]
(*@\textbf{User: }@*)
The uploaded images are consecutive frames from a video. The numbers in the file name indicate the order of the frames in the sequence, so frame_1.jpg is the first frame, followed by frame_2.jpg which is the second frame, etc. The sequence shows an interaction between a hand and an object. An interaction usually begins when an object starts to move with the hand. An interaction usually ends when the hand starts to move without the object. 
Q: In which frame does the interaction end? 
Answer briefly with: "Prediction: <frame number>" 
\end{lstlisting}
\end{figure}

\begin{figure}[t]
\begin{lstlisting}[style=prompt, basicstyle=\scriptsize\ttfamily, escapeinside={(*@}{@*)}, caption={A prompt used in the OS experiment.}, label={lst:os_prompt}]
(*@\textbf{System:}@*) 
You are ...
(*@\textbf{Images:}@*) 
[First frame of example video] 
...
[Last frame of example video]
(*@\textbf{User: }@*)
The uploaded imgaes ... 
Q: In which frame does the interaction end? 
Answer briefly with: "Prediction: <frame number>" 
A: Prediction: 3
(*@\textbf{Images:}@*) 
[First test video frame] 
...
[Last test video frame]
(*@\textbf{User: }@*)
The uploaded images ...
Q: In which frame does the interaction end? 
Answer briefly with: "Prediction: <frame number>" 
\end{lstlisting}
\end{figure}

\vspace{-1em}
\paragraph{One-Shot (OS) regime.} In the OS regime, models are provided with a single example of the main task prior to performing the task on the test video sequence. The example, includes a video with a different event from the experimental dataset and the annotated true frame for this video, where the core event occurs. The prompt is provided in~\cref{lst:os_prompt}. We evaluate the models on each test video in a leave-one-out approach, and average the accuracy for that video across all trials. 

\vspace{-1em}
\paragraph{Two-Shot (TS) regime.} We further extended the experiment of the OS regime, to the TS regime, by presenting to the model a second example of another event with the corresponding labeled frame, prior to the test. As in the OS regime, we average the accuracy across many trials for each test video (same total trials as in the OS regime), where in each trial the two examples are drawn randomly from the set of experimental dataset, excluding the test video. 

\vspace{-1em}
\paragraph{\textit{Grounding} condition.} The grounding procedure precedes the main test task. The model is instructed to describe the contents of the provided image sequence, and to repeat a particular piece of information from the instructing prompt. The motivation behind this procedure came from a preliminary experiment that was conducted with the online User-Interface of ChatGPT-4o, in which the model had better understanding of the instructing prompt and more attention to the image contents, when explicitly was asked about them. The prompt used for the grounding condition is shown in \cref{lst:gnd_prompt}. 

\begin{figure}[t]
\begin{lstlisting}[style=prompt, basicstyle=\scriptsize\ttfamily, escapeinside={(*@}{@*)}, caption={A grounding prompt in the OS experiment to verify model's understanding of both verbal and visual contents of a provided example.}, label={lst:gnd_prompt}]
(*@\textbf{System:}@*) 
You are ...
(*@\textbf{Images:}@*) 
[First example video frame] 
...
[Last example video frame]
(*@\textbf{User: }@*)
The uploaded imgaes ... 
To check your understanding, please repeat the correct answer to the example question, and specify which object was in contact with the hand in these frames? Answer with the object's name.
\end{lstlisting}
\end{figure}

\begin{figure}[t]
\begin{lstlisting}[style=prompt, basicstyle=\scriptsize\ttfamily, escapeinside={(*@}{@*)}, caption={An instructing prompt for step-by-step reasoning.}, label={lst:rsn_prompt}]
(*@\textbf{System:}@*) 
You are ...
(*@\textbf{Images:}@*) 
[First test video frame] 
...
[Last test video frame]
(*@\textbf{User: }@*)
The uploaded images ...
Q: In which frame ... ?
Answer briefly with ... 
Think step by step, and show all intermediate reasoning before giving the final answer.
\end{lstlisting}
\end{figure}

\begin{figure*}[t]
    \centering
        \includegraphics[width=0.95\textwidth]{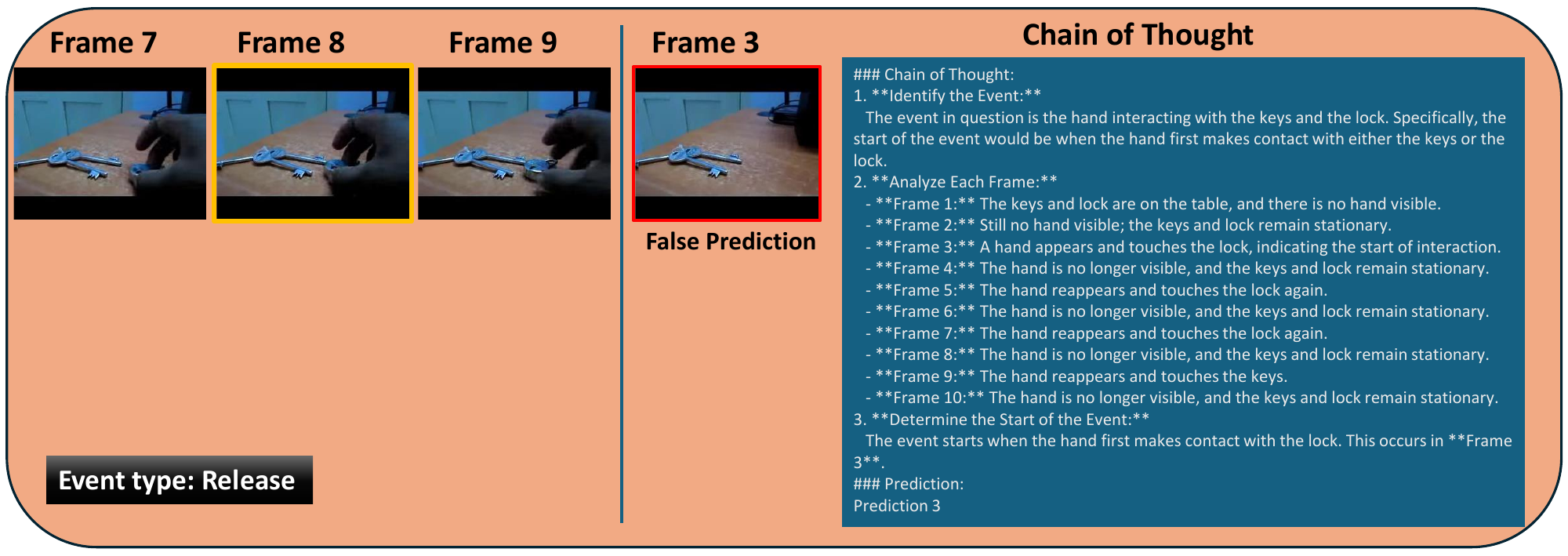}
    \caption{\textbf{Example false prediction.} The model provides the chain-of-thought under the ``WITH'' \textit{Reasoning} condition. The example shows that the reasoning seems logical and realistic, but the grounding of the physical release event in the video frames is very loose. The orange box marks the true frame. The red box marks the false prediction. }
    \label{fig:example_predictions}
\end{figure*}

\vspace{-1em}
\paragraph{\textit{Reasoning} condition.} The reasoning procedure is combined with the main task, by instructing the model to describe in detail the reasoning behind its answer to the main frame detection task (see prompt in \cref{lst:rsn_prompt}). This prompt does not modify the internal reasoning parameters of the model. The motivation for this experimental condition comes from recent studies showing that models' performance increases when they are explicitly instructed to provide a step-by-step description, and thus may extract and combine more relevant information in the final answer \cite{wei_cot_prompting2022}. \cref{fig:example_predictions} shows an example of a false prediction together with the model's step-by-step chain of thought. 

\section{Results}
\label{sec:results}
In this section we describe in detail the results from the experiments conducted in this study.

\paragraph{Event frame detection.} The main task for the models was to detect the frame where a test event ('contact' or 'release) occurs. A correct detection is considered within an allowed error tolerance $e_\tau$. For example, $e_\tau=0$ means the model predicted the exact true (human) labeled frame where the event occurs, and $e_\tau=1$ means that the predicted frame can be up to one-frame off (before or after) the true frame. 
\setlength{\tabcolsep}{3pt}
\begin{table}[b]
    \footnotesize
    \caption{\textbf{Reasoning effect on LMMs' performance.} Models are instructed to describe step-by-step the reasoning behind their prediction of the frame where the interaction occurs. Mean accuracy is measured for the detection of the frame where the test event occurs, within an allowed error tolerance (here the exact or one frame off the true frame). Results are reported as per 3 ICL regimes (ZS, OS, TS) for the models: (i) Gemini-2.5-Flash and (ii) GPT-5.2.}
    \centering
            \begin{tabularx}{1.0\linewidth}{l|c|cc|cc}
            \toprule
                \textbf{ICL}                 & \textbf{RSN} & \multicolumn{4}{c}{\textbf{Mean Accuracy Percentage (SD)}} \\ 
                & & \multicolumn{2}{c|}{\textbf{Exact}} & \multicolumn{2}{c}{\textbf{1-off}} \\
                & & \multicolumn{1}{c}{\textbf{Gemini}} & \multicolumn{1}{c|}{\textbf{GPT}} & \multicolumn{1}{c}{\textbf{Gemini}} & \multicolumn{1}{c}{\textbf{GPT}} \\
                \midrule
                \multirow{2}*{\textbf{ZS}}  & W/O  & $8.08$ $(1.94)$ & $9.60$ $(2.10)$ & $30.81$ $(3.29)$ & $\textbf{35.35}$ $(3.41)$ \\
                & W  & $9.09$ $(2.05)$ & $\textbf{12.12}$ $(2.33)$  & $29.80$ $(3.26)$  & $33.84$ $(3.37)$ \\
                \midrule
                \multirow{2}*{\textbf{OS}}    & W/O   & $\textbf{19.87}$ $(2.84)$ & $18.15$ $(2.29)$ & $\textbf{47.15}$ $(3.55)$ & $42.96$ $(3.19)$ \\
                & W  &$16.74$ $(2.65)$ & $15.93$ $(2.02)$  & $43.71$ $(3.53)$ & $39.43$ $(3.05)$ \\
                \midrule
                \multirow{2}*{\textbf{TS}} & W/O & $\textbf{20.71}$ $(20.71)$ & $18.03$ $(2.73)$  & $\textbf{49.07}$ $(3.55)$ & $44.31$ $(3.53)$ \\
                & W   & $17.98$ $(2.73)$ & $17.12$ $(2.68)$ & $47.53$ $(3.55)$ & $40.82$ $(3.49)$ \\
                \bottomrule
            \end{tabularx}
    \label{tab:results_rsn}
\end{table}
As demonstrated in~\cref{fig:example_events}, it is often evident that there is no physical contact between the hand and target object already one frame after a release or before a contact event. Throughout this section, the models' performance is measured as the mean detection accuracy (with standard error) across all test events (99) in the experimental dataset.

\setlength{\tabcolsep}{3pt}
\begin{table}[b]
    \footnotesize
    \caption{\textbf{Grounding effect on LMMs' performance.} Prior to the frame detection task, models are instructed to name the target object, to improve their perceptual grounding. Mean accuracy is measured for the detection of the frame where the test event occurs, within an allowed error tolerance (here the exact or one frame off the true frame). Results are reported as per 3 ICL regimes (ZS, OS, TS) for the models: (i) Gemini-2.5-Flash and (ii) GPT-5.2.}
    \centering
            \begin{tabularx}{1.0\linewidth}{l|c|cc|cc}
            \toprule
                \textbf{ICL}                 & \textbf{GRND} & \multicolumn{4}{c}{\textbf{Mean Accuracy Percentage (SD)}} \\
                & & \multicolumn{2}{c|}{\textbf{Exact}} & \multicolumn{2}{c}{\textbf{1-off}} \\
                & & \multicolumn{1}{c}{\textbf{Gemini}} & \multicolumn{1}{c|}{\textbf{GPT}} & \multicolumn{1}{c}{\textbf{Gemini}} & \multicolumn{1}{c}{\textbf{GPT}} \\
                \midrule
                \multirow{2}*{\textbf{ZS}}   & W/O  & $7.58$ $(1.89)$ & $\textbf{11.62}$ $(2.28)$ & $29.29$ $(3.24)$ & $\textbf{34.85}$ $(3.39)$  \\
                & W  & $9.60$ $(2.10)$ & $10.10$ $(2.15)$  & $31.31$ $(3.30)$  & $34.34$ $(3.38)$ \\
                \midrule
                \multirow{2}*{\textbf{OS}}    & W/O & $17.78$ $(3.84)$ & $17.31$ $(2.18)$  & $44.77$ $(3.53)$ & $41.94$ $(3.16)$ \\
                & W  & $\textbf{18.84}$ $(2.78)$ & $16.77$ $(2.14)$   & $\textbf{46.09}$ $(3.54)$ & $40.45$ $(3.09)$ \\
                \midrule
                \multirow{2}*{\textbf{TS}} & W/O & $19.32$ $(2.81)$ & $19.31$ $(2.81)$ & $\textbf{49.49}$ $(3.55)$ & $45.22$ $(3.54)$ \\
                & W & $\textbf{19.37}$ $(2.81)$ & $15.84$ $(2.60)$  & $47.10$ $(3.55)$ & $39.92$ $(3.48)$ \\
                \bottomrule
            \end{tabularx}
    \label{tab:results_gnd}
\end{table}

\begin{figure*}[htbp]
    \centering
    \begin{subfigure}{0.35\textwidth}
        \centering
        \caption{Mean Accuracy (ZS)}
        \includegraphics[width=1\textwidth]{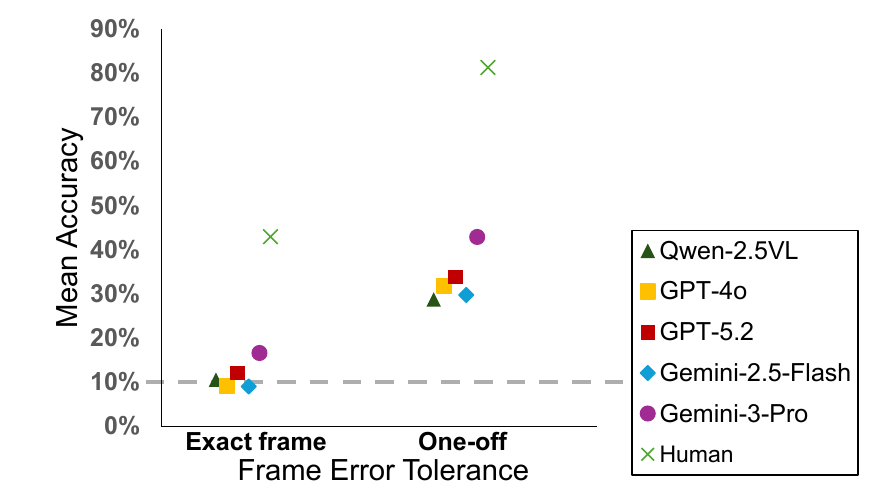}
    \end{subfigure}%
    \begin{subfigure}{0.35\textwidth}
        \centering
        \caption{Mean Frame Error (ZS)}
        \includegraphics[width=1\textwidth]{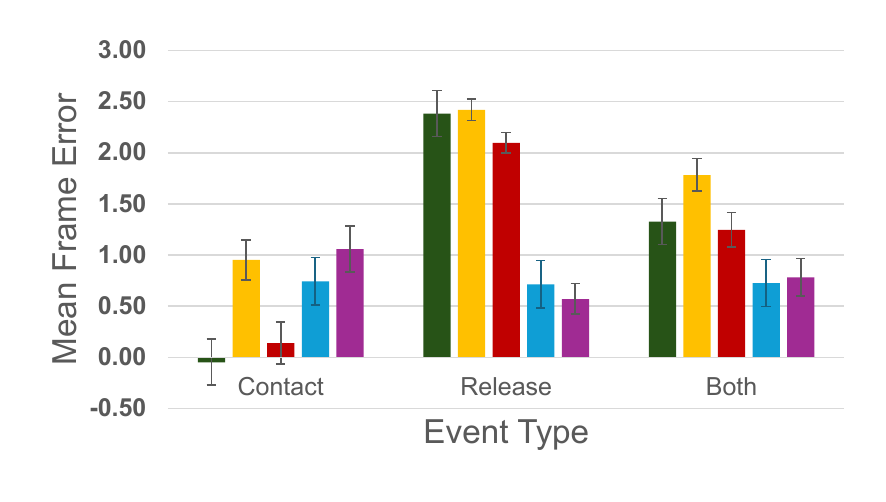}
    \end{subfigure}
    \begin{subfigure}{0.295\textwidth}
        \centering
        \caption{GPT-5.2}
        \includegraphics[width=1\textwidth]{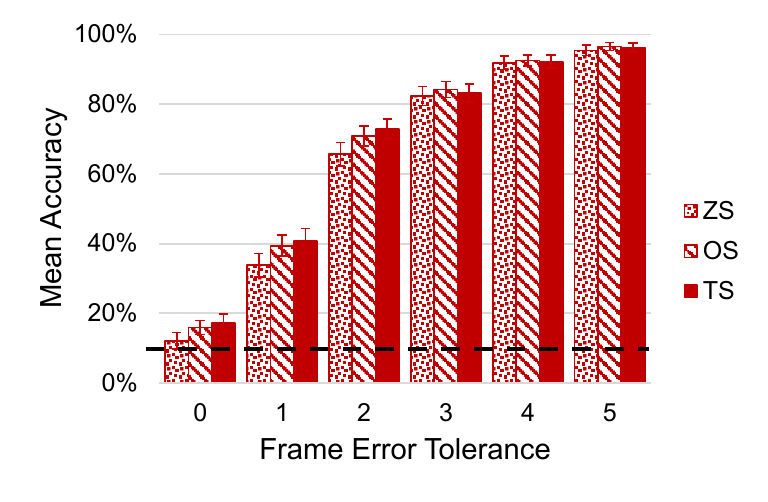}
    \end{subfigure}
    \caption{\textbf{Models' performance.} (a) A comparison between models' performance at Zero-Shot ICL. Exact detection of the physical event is around chance level for most models. At one-frame off error tolerance Gemini-pro-3 peeks at 42.9\%. Humans outperform the models by far. (b) The models' mean frame error. A positive error indicates a prediction which precedes a physical contact event, or follows a physical release event. In these frames there is often a clear gap between the hand and the target object. (c) Detail mean accuracy vs. frame error tolerance for GPT-5.2. The chart depicts a slight performance increase across the different ICL regimes (ZS, OS, TS). Results are under the ``WITH'' \textit{Reasoning} condition. Chance level is 10\% (shown as a dashed horizontal line) as the length of all videos in the experimental dataset is 10 frames.}
    \label{fig:acc_vs_errtol}
\end{figure*}

\cref{fig:acc_vs_errtol} presents performance charts for the evaluated models. The results in \cref{fig:acc_vs_errtol}a indicate that at the ZS regime, the models' accuracy is about chance level (10\%) for detecting the exact frame where the event occurs. The accuracy for an error tolerance of one-frame off peaks at 42.9\% with Gemini-3-Pro. However, as indicated by \cref{fig:acc_vs_errtol}b, the models' failure often tends to frames with no visual grounding of a physical contact. In addition, when evaluated on 'non-event' control videos, which contained hands and objects but did not present a physical contact, most of the models completely failed. Except for Qwen, which responded correctly at 18.2\% of the trials that there is no event in these video, the rest of the models always predicted a frame and provided an explanation for their choice. The chart in \cref{fig:acc_vs_errtol}c depicts the GPT-5.2 model's performance as a function of the allowed error tolerance (other models yielded similar results, see Supplementary). The chart also shows a minor increase in performance between the different ICL regimes as was demonstrated already in earlier studies \cite{dong_icl_survey2024}. 

\cref{tab:results_rsn} reports the mean accuracy for $e_\tau=0$ and $e_\tau=1$ as per each of the three ICL regimes (ZS, OS, TS) under the \textit{Reasoning} condition (see \cref{sec:experiments}).
Similarly, \cref{tab:results_gnd} lists the mean accuracy under the \textit{Grounding} condition (see supplementary for the results on other models).

These results indicate that explicitly instructing the models to provide visual grounding evidence and reasoning for their answers does not necessarily increase their performance. Surprisingly, the results indicate that these instructions, may sometimes yield a decrease in the models' performance. This reflects the foundational gap between System 1 and System 2 abilities of the models. We further discuss this in \cref{sec:discussion}.

Finally, we performed an ablation experiment, inspired by recent work on boosting LLMs reasoning with a verification feedback procedure -- similar to reinforcement learning approach \cite{wei_cot_prompting2022, yao_tot2023, guo_deepseekr12025} -- but this did not improve the models' performance on the detection task (see supplementary). 

\vspace{-1em}
\paragraph{Action and object recognition.} 
\setlength{\tabcolsep}{3pt}
To contextualize the failure in fine-grained grounding, we benchmarked the models’ high-level action and object recognition capabilities. Using our experimental dataset consisting of a subset of 33 videos across 15 action templates from SSv2, we evaluated two distinct tasks:

(i) Action Classification: Models were prompted to identify the correct SSv2 action template for each video. To ensure evaluation precision and avoid linguistic ambiguity, we required models to output numerical template IDs rather than text strings. Accuracy was measured using Top-1 and Top-5 metrics. For few-shot guidance, the prompt included verbal descriptions of four representative interaction styles, excluding actual video frames.

(ii) Object Naming: Given the correct action template, models were tasked with replacing "something" placeholders with the specific objects identified in the video. Accuracy was determined by manually verifying predicted objects against ground-truth labels, accounting for ordering and synonyms.

\begin{table}[b]
\footnotesize
\begin{tabularx}{1.0\linewidth}{lccc}
\toprule
& \multicolumn{2}{c}{\textbf{Action Classification Acc.} (\%)} & \textbf{Obj. Naming Acc.} (\%) \\
\textbf{Model} & \textbf{Top-1} & \textbf{Top-5}  \\
\midrule
GPT-4o   & $48.48\pm8.69$ & $81.82\pm6.71$ & $81.82\pm6.71$\\
GPT-5.2  & $45.45\pm8.67$  & $\textbf{96.97}\pm2.98$ & $93.94\pm4.15$\\
Gemini-2.5  & $\textbf{66.67}\pm8.33$  & $90.91\pm5.08$ & $93.94\pm4.15$ \\
Gemini-3-Pro & $63.64\pm8.37$  & $87.88\pm5.68$ & $\textbf{96.97}\pm2.98$\\
Qwen-2.5VL  & $60.61\pm8.51$  &  $84.85\pm6.24$ & $69.70\pm8.00$\\

\bottomrule
\end{tabularx}
\caption{Per-model accuracy statistics for global action classification and naming the correct target objects in each test video.}
\label{tab:acc-class-object}
\end{table}

As shown in \cref{tab:acc-class-object}, the models demonstrate robust global understanding, with most achieving over 85\% Top-5 action accuracy and high object naming precision. This confirms that their struggle to pinpoint interaction events is not due to a lack of semantic or object-level recognition, but rather a specific deficit in spatiotemporal grounding.

%
\vspace{-1em}
\paragraph{Event bounding-box (BBox) detection.} We evaluated the models ability to locate the interaction region within the predicted video frame, where the event occurs. Since the annotations contain a point image location per event (point of contact or release), we considered for the evaluation several formulations for the prediction of the event location, including exact point location and different BBox formats. We report here the best performing setting, which was predicting a BBox, and measuring the Intersection-over-Union (IoU) between the prediction and a square BBox centered around the label point location.
We tested several box sizes in the range 20 to 200 pixels. Using a 120$\times$120 pixel size boxes around the annotation point yielded the best performance: mean IoU of only $0.147$ for Gemini-3-Pro and $0.086$ for GPT-5.2 (see supplementary).  
Overall, these findings indicate that the model is unable to reliably localize the regions around the physical interaction.
\vspace{-1em}
\paragraph{Human performance.} As a comparison to the models performance, we asked two naive humans to perform the same tasks on the experimental dataset. For humans, the mean accuracy on the event frame detection task was $43.00\%\pm5.04\%$ for the exact frame, and $81.36\%\pm4.18\%$ for 1-frame off - high above the best performing model as shown in Fig.~\ref{fig:acc_vs_errtol}a. The ICC between the human subjects was $0.87$. The accuracy on the 'non-event' control set was $70.26\%\pm6.93\%$. The mean IoU for the event location detection was $0.71\pm0.02$.

\section{Discussion}
\label{sec:discussion}

Recent advancement of current vision models improve significantly LMMs generalization ability in recognizing unseen actions and scenes in real-world videos \cite{tang_llm_video_survey2025, buschoff_nature2025, tong2022videomae}. Our action recognition evaluation verify this increased performance for five models: Qwen-2.5VL-72B, GPT-4o, GPT-5.2, Gemini-2.5-Flash and Gemini-3-Pro (see~\cref{tab:acc-class-object}).
However, previous studies have indicated a possible limitation of current vision models in understanding the core interaction events underlying the general action \cite{Gunawardhana2024SelfSupervised}. Do LMMs overcome this limitation by leveraging their vast common knowledge and visual semantics? 

In this study we conducted a series of experiments, under several ICL regimes, to test LMMs' ability in detecting where and when in the video, core interaction events occur. Specifically, we focused on `contact' and `release' events, where a target object becomes attached to an agent (e.g., a hand) or detached from the agent (\cref{fig:example_events}; \cref{sec:experiments}). We introduced the \textit{Contact-Release Interaction Dataset (CRID)} -- a new large scale dataset with more than 20K human annotated events in videos from the SSv2 dataset \cite{goyal2017something} (\cref{sec:dataset}).

Our experimental results indicate that despite the System 1 ability of the models to classify correctly the action in the videos ($>80\%$ in Top-5) and even name the correct target objects in the scenes ($>70\%$), the models struggle with detecting the physical core events and ground them visually in the videos ($<17\%$). Introducing similar examples using few-shot ICL paradigm slightly improves the performance, which still remains slightly above chance level (\cref{fig:acc_vs_errtol}c). 

In contrast with earlier studies \cite{wei_cot_prompting2022, shao_visual_cot2024}, applying Chain-of-Thought prompting does not necessarily increase the models' performance (see \cref{tab:results_rsn}). Similarly, explicitly instructing the models to attend and describe the input video (e.g., name the target objects in the interaction scene), does not improve the models' grounding ability and hence the performance (see \cref{tab:results_gnd}). These results suggest that the models lack System 2 understanding abilities of core interaction events. 

We find that models struggle with the perceptual grounding of the core events underlying actions and interactions in the visual input, despite their general ability to describe the action and participating objects and agents in the interactions. This limitation is partly related to the challenge of complex question decomposition as was already shown in previous studies \cite{zhang_vqa_dec2024}. However, it seems that there is more to this limitation. We hypothesize that the main limitation is rooted in a loose integration between the visual representation (often of pretrained visual transformers) and the language representation, which are mostly trained separately. This limitation projects also to the models' inability to overcome current challenges of visual models in interpreting spatial relations between objects \cite{liu-etal-2025-multimodal-large} and complex dynamic events, despite their huge semantic knowledge. In a sense, the models exhibit a ``shortcut learning'' behavior and merely able to tell a ``good story'' about possible interactions when hands appear in proximity of objects in scenes. In struggling to pinpoint the moment and location of physical contact that defines the interactions, the models lack the perceptual grounding required for deeper understanding of dynamic scenes

The implication of this limitation may be that current LMMs lack the capacity to develop full visual understanding of dynamic interactions, similar to intelligent beings \cite{michotte1963perception, buschoff_nature2025}, and thus can have only limited ability in interpreting unfamiliar and complex interactions, as well as in planning interactions on their own for artificial systems.

\section{Conclusions}
\label{sec:conclusions}
In this paper we demonstrate a major limitation of current large multi-modal models in understanding dynamic interactions. 
Our analysis suggests that current models are operating as sophisticated System 1 engines. They recognize a `picking up' action by the presence of a hand and a cup, but they do not perform the System 2 `mental simulation' required to pinpoint the exact moment of physical attachment. To move toward genuine multi-modal intelligence, future architectures must incorporate structured priors or causal world models that treat interaction events not just as pixels, but as discrete physical state changes (e.g., attend to motion and motion boundaries around the hand and the object). We introduce CRID - an extension to the SSv2 dataset with more than $20K$ detailed annotations of core physical events in more than $10K$ videos. These annotations may be used in future efforts to develop new architectures or foundation models with cognitive understanding of visual dynamic interactions.

\clearpage
{
    \small
    \bibliographystyle{ieeenat_fullname}
    \bibliography{main}
}

\pagebreak
\setcounter{page}{1}
\renewcommand{\thetable}{S\arabic{table}}
\renewcommand{\thefigure}{S\arabic{figure}}
\newcommand{\sfigref}[1]{Fig.~\ref{#1}}
\newcommand{\stabref}[1]{Tab.~\ref{#1}}
\renewcommand{\thelstlisting}{S\arabic{lstlisting}}
\newcommand{\slistref}[1]{Listing~\ref{#1}}
\setcounter{table}{0}
\setcounter{figure}{0}
\setcounter{lstlisting}{0}

\maketitlesupplementary

\section{Ablation Experiments}
\label{sec:supp_exp_detail}

\subsection{Action and object recognition}

This section provides extended descriptions of the experimental setup of the evaluation on the tasks related to action and object recognition, mentioned in Section~4 of the main manuscript. All experiments were conducted using the extracted video frames of the full length videos from the original SSv2 dataset \cite{goyal2017something}. The models were given, as input, the sequence of video frames and prompted with the task-specific textual information detailed below. Five models including, GPT-4o, GPT-5.2, Gemini-2.5-Flash, Gemini-3-Pro and Qwen-2.5-VL-72B, were evaluated under the zero-shot regime (without any examples), and without any of the modifying conditions (\textit{Grounding} and \textit{Reasoning}).



Task-specific textual information was provided only when required:
\begin{itemize}
    \item \textbf{Action-template recognition:}  
    The prompts included the list of candidate template labels and their corresponding template IDs.
    
    \item \textbf{Object-placeholder extraction:}  
    The prompts included the template ID, the template sentence, and the number of placeholder slots required by the template.
    
    \item \textbf{Event bounding-box detection:}  
    No additional textual information was provided beyond the sequence of video frames.
\end{itemize}

\paragraph{Action and object recognition.} \slistref{supplst:placeholder_prompt} and \cref{supplst:act_recog_prompt} presents the prompts used in the experiments for testing the models performance in the tasks of object and action recognition, respectively. 
In the listed prompts, the strings \texttt{template\_sentence}, 
\texttt{template\_id}, and \texttt{n\_slots} denote variables that were automatically replaced during execution for each experiment, according to the specific interaction template. The prompt structure itself was fixed, while these fields changed to reflect the corresponding template text, its numeric ID, and the number of placeholder slots.

\paragraph{Event bounding-box detection.} \cref{supplst:bbox_prompt} present the prompt used in the experiment for testing the models performance in the task of detecting the spatial location where the event occurs in the predicted frame.

\subsection{Tow-Shot with Feedback}
Inspired by recent work on boosting LLMs reasoning with verification feedback \cite{wei_cot_prompting2022, yao_tot2023, guo_deepseekr12025}, we designed a variant to the common TS regime, in which the label of the second example was presented to the model indirectly, through an iterative feedback session. In this session, the model performed the detection task on the second example video and then prompted a numerical verification feedback by the user side. The feedback indicate a metric on the gap between the predicted frame ($f_p$) and the true frame ($f_t$) of the event. We defined the error in prediction, $\epsilon(f_t, f_p)$, via a sigmoid function, as shown in \cref{eq:error}. The error function is shifted to the middle of the frame range ($\frac{N}{2}$) (as we constrained the error to be in the range of $[0, 1]$, and required it to be $\epsilon(\frac{N}{2}) = 0.5$). The score was defined as shown in \cref{eq:score}.
The iterative feedback session ended when the model predicted the correct frame, or after it exceeded a limit of allowed trials $T_{th}$, which in our experiments was set to 10 (equivalent to the maximal number of iterations in the naive case where the model simply scans all the frames in turn until it gets to the right frame). After the iterative session ended, the model was instructed to perform the main detection task on the test video. 
In our experiment, we included examples only of other events from the same full video from which the test event was cropped, thus providing the model context with familiar context from the test video. 
The protocol followed the algorithm in \cref{alg:feedback}. The instructing prompt is presented in \slistref{lst:ts_fb_prompt}.

\begin{equation}
    \begin{split}
        \epsilon(f_t, f_p) = \sigma(|f_t - f_p| + \frac{N}{2}) = \frac{1}{1+e^{-|f_t-f_p| + \frac{N}{2}}}
    \end{split}
    \label{eq:error}
\end{equation}

\begin{equation}
    \begin{split}
        s(f_t, f_p) = 1 - \epsilon(f_t, f_p) 
    \end{split}
    \label{eq:score}
\end{equation}

\begin{algorithm}[t]
\footnotesize
\caption{Iterative feedback algorithm}
\label{alg:feedback}
\begin{algorithmic}
\Ensure $fbScr = 0.5$
\Ensure $prvPred_0 = 0$
\Ensure $prvPred_1 = 0$
\While{$0.98 > |fbScr| $}
    \If{$(sgn(fbScr) > 0) \And (prvPred_1 \geq prvPred_0)$}    
        \State $curPred \gets [prvPred_1, 10]$  
    \ElsIf {$(sgn(fbScr) > 0) \And (prvPred_1 \le prvPred_0)$}
        \State $curPred \gets [1, prvPred_1]$
    \ElsIf {$(sgn(fbScr) < 0) \And (prvPred_1 \geq prvPred_0)$}
        \State $curPred \gets [prvPred_0, prvPred_1]$
    \ElsIf {$(sgn(fbScr) < 0) \And (prvPred_1 \le prvPred_0)$}
        \State $curPred \gets [prvPred_1, prvPred_0]$
    \EndIf
    \State $prvPred_0 \gets prvPred_1$
    \State $prvPred_1 \gets curPred$
    \State $fbScr \gets MeasureScore(prvPred_0, prvPred_1, trueFrame)$
\EndWhile
\end{algorithmic}
\end{algorithm}

\begin{figure}[b]
\begin{lstlisting}[
    float=false,
    style=prompt,
    basicstyle=\scriptsize\ttfamily,
    escapeinside={(*@}{@*)},
    caption={Instruction prompt for object detection.},
    label={supplst:placeholder_prompt}
]
(*@\textbf{System:}@*) 
You are an expert video-interaction classifier.
(*@\textbf{User:}@*) 
The uploaded images are consecutive frames from a video. The numbers in the file name indicate the order of the frames in the sequence, so frame_0.jpg is the first frame, followed by frame_1.jpg which is the second frame, etc.
You will see frames from a short video and one template sentence.
Template (with placeholders):
{template_sentence}
Return JSON ONLY (no extra text, no markdown fences):
{
  "template_id": {template_id},
  "placeholders": [
    "<slot1>",
    "<slot2>",
    ...
  ]
}
Rules:
- Provide exactly {n_slots} placeholders, ordered left-to-right as they appear in the template.
- Use short, concrete noun phrases for visible objects (e.g., "potato", "vicks vaporub bottle").
- Avoid generic words such as "object", "thing", or "item".
- Do not include explanations, labels, confidence scores, or any additional fields.
\end{lstlisting}
\end{figure}

\begin{figure}[t]
\begin{lstlisting}[float=false, style=prompt, basicstyle=\scriptsize\ttfamily, escapeinside={(*@}{@*)}, caption={Instruction prompt for event bounding-box detection.}, label={supplst:bbox_prompt}]
(*@\textbf{System:}@*)
You are a useful assistant and an expert in video understanding.
(*@\textbf{User:}@*)
The uploaded images are consecutive frames from a video. The numbers in the file name
indicate the order of the frames in the sequence, so frame_1.jpg is the first frame,
followed by frame_2.jpg which is the second frame, etc. The sequence shows an interaction between a hand and an object. An interaction usually begins when an object starts to move with the hand. An interaction usually ends when the hand starts to move without the object.
Q: In which frame does the interaction end?
Answer briefly with: 'Predicted frame: <frame number only>'
Q: Where in the predicted frame does the interaction occur?
Answer with bounding box coordinates in the original frame's pixel scale:
'Predicted BBox: [<left_x>, <top_y>, <width>, <height>]'
Bounding box requirements:
- Use the SAME coordinate scale as the original video frame (pixels, not normalized).
- The box should tightly enclose the MAIN CONTACT REGION between the interacting objects (e.g., where a hand touches an object), not the entire objects.
- Make the bounding box as SMALL as possible while still fully containing this contact region.
- The bounding box MUST stay within the frame boundaries; never extend beyond the image edges.
- If uncertain, err on slightly smaller rather than larger, as long as the contact area is included.
\end{lstlisting}
\end{figure}

\begin{figure*}[t]
\begin{minipage}{0.95\textwidth} 
\begin{lstlisting}[float=false, style=prompt, basicstyle=\scriptsize\ttfamily, escapeinside={(*@}{@*)}, caption={Instruction prompt for action recognition.}, label={supplst:act_recog_prompt}]
(*@\textbf{System:}@*) 
You are an expert video-interaction classifier.
(*@\textbf{User:}@*) 
The uploaded images are consecutive frames from a video. The numbers in the file name indicate the order of the frames in the sequence, so frame_0.jpg is the first frame, followed by frame_1.jpg which is the second frame, etc.
You will see frames from a short video.
Choose the FIVE best-matching interaction templates (ranked by confidence).
Your goal:
Pick the template ID corresponding to the template label that best describes what happens in the sequence of frames. Focus on the physical interaction between visible objects.
Illustrative Examples (for clarity):
- If the frames show a human hand placing several books one after another on a shelf, the correct template is "Putting number of something onto something" - because the action repeats multiple times and involves a series of objects being placed on another object (the shelf).
- If the frames show a person putting three distinct objects on a table, the correct template is "Putting something, something and something on the table" 
- because exactly three objects are placed on the surface at once.
- If the frames show one object being placed next to another, the correct template is "Putting something next to something".
- If the frames show an object being placed inside another, the correct template is "Putting something into something".
Return JSON ONLY (no text or markdown fences):
[
    { "template_id": <int> },
    { "template_id": <int> },
    { "template_id": <int> },
    { "template_id": <int> },
    { "template_id": <int> }
]
Rules:
- Return exactly 5 objects, ranked most->least confident.
- Each object MUST have only one field: "template_id" (integer from the list below).
- Do NOT include any text, explanations, or reasoning.
- Choose IDs based purely on what the video depicts.
- Prefer the main **physical interaction** over camera motion.
Disambiguation:
- Return exactly 5 objects, ranked most->least confident.
- "into" -> containment / inside relation.
- "onto" -> on top of.
- "next to" -> lateral adjacency without contact stacking.
- "slanted surface" -> object accelerates along a plane.
- "on a flat surface w/o rolling" -> stable placement without motion.
Scoring & Specificity Rules (very important):
- Prefer the template whose action AND number of involved objects best match the scene.
- If THREE distinct objects interact, prefer a 3-object template over any 2-object option, if the action fits.
- Tie-breakers: object count > verb precision > physical outcome > surface/slant qualifiers.
- Don't pick a broader template if a more specific one fits.
Templates (id: label):
    1: Attaching something to something
    33: Moving part of something
    48: Piling something up
    54: Poking a stack of something without the stack collapsing
    57: Poking something so that it falls over
    58: Poking something so that it spins around
    88: Pulling something onto something
    97: Pushing something so it spins
    98: Pushing something so that it almost falls off but doesn't
    99: Pushing something so that it falls off the table
    102: Putting number of something onto something
    120: Putting something, something and something on the table
    122: Rolling something on a flat surface
    144: Stacking number of something
    148: Taking something out of something
\end{lstlisting}
\end{minipage}
\end{figure*}

\subsection{Chain-of-Thought (CoT) Tuning}
In the main experiment, examples included only videos with the true frame, but without further description of the dynamic event. In contrast, in this experiment we provided the agent with detailed steps for detecting the frame where the event occurs, for each example video. Taking inspiration from \cite{wei_cot_prompting2022}, in which the authors showed that an LMM can improve in the task of solving mathematical problems by presenting it examples of solutions to similar mathematical problems with a detailed, step-by-step solutions. 

In our experiments we fed the agent a prompt that included three parts, viz. 1) a general explanation of the goal the agent was required to achieve, 2) a set of 1 --- 8 examples with a full solution, and 3) the final test set, for which the final accuracy was computed (see \cref{lst:cot-exp}). For this matter we manually formulated 18 CoT prompts (9 for contact and 9 for release events), each with a detailed explanation of the scene and the interactions between a hand and objects located in it. 

\begin{figure}[h]
\begin{lstlisting}[style=prompt, basicstyle=\scriptsize\ttfamily, escapeinside={(*@}{@*)}, caption={Instruction prompt for TS with feedback experiment.}, label={lst:ts_fb_prompt}]
(*@\textbf{User: }@*)
In the next procedure follow these rules: 
1) Your output should ALWAYS be the word "Prediction" followed by the frame number.
2) You will be provided a score with absolute values in the range [0, 1], representing the correctness of your prediction.
3) Score of 0 means that your prediction is incorrect, while an absolute value of 1 means you have found the correct frame.
4) The absolute values in the range of [0,1] reflect the proximity of the predicted frame to the true target frame, where the higher value is better.
5) The sign of the score signifies the direction for your next prediction. If the sign is positive, your next prediction should be in the same direction as the previous prediction. A negative sign means you should change the direction of your next prediction. For example, if your last prediction was frame 5, your current prediction is frame 6 and the score is negative - your next prediction should be smaller than 5. On the other hand, if the last prediction was frame 7, the current is frame 4 and your score is negative - you should predict values greater than 4. 
6) You should never predict the same frame twice.
7) Stop your predictions only when the score is above 0.98.
Now we will perform an iterative session, during which you will need to find the frame in the provided image sequence. Follow the instructions above when prompted with the feedback score.
\end{lstlisting}
\end{figure}

\begin{figure}[b]
\begin{lstlisting}[style=prompt, basicstyle=\scriptsize\ttfamily, escapeinside={(*@}{@*)}, caption={The prompt template provided to the LMM agent in course of the CoT prompt tunning experiment. The prompt included an introduction describing a general goal the agent was required to achieve; a set of 1 to 8 examples, each including a detailed CoT, which described the objects in the scene and the steps towards the correct solution, and the corresponding frames of the examples; finaly, the test sequence was included as the last part of the prompt.}, label={lst:cot-exp}]
(*@\textbf{User: }@*)
You are an expert in video understanding and motion analysis.

We are interested in detecting and interpreting interactions in video sequences. We focus on interactions between hands and objects.
An interaction usually begins with a contact between a hand and a target object, when the static target object starts to move with a hand holding it.
An interaction usually ends when a hand releases a target object and starts to move without the object after it ungrasps the object.

You will be provided with consecutive frames from video sequences, where the order of the frames in each sequence is deteremined by the upload ordering and the acending index numbers in the file names.
You are asked to detect the exact frame where the interaction occurs. Your final answer should be brief in the format: "Prediction: <frame number>". Before your final answer, provide a step by step detailed description of all intermediate reasoning grounded in the video frames, in support of your final answer.

Example1:

<example video frame #1>
...
<example video frame #10>

STEPS:
1. Identify the Event:
   < example text >
2.  Frames analysis:
	< example text >
3. Conclusion for prediction:
   < example text >.
4. Final answer:
	Prediction: < example true frame >
	
Example2: 
...

Example8:
...

Test:
The following is the test video sequence. 
<test video frame #1>
...
<test video frame #10>

Please detect to exact frame where the interaction occurs. Similar to the examples, provide a detailed reasoning before your final answer.
\end{lstlisting}
\end{figure}


As before, at each iteration we requested the LMM agent to detect the exact frame in which the interaction (i.e., contact or release) happens, but this time, we provided it also with a subset of examples, with a varying length between 1 to 8, similarly to the experiments in \cite{wei_cot_prompting2022}. In each iteration a random set of examples (of the same type as the test video) were chosen and their videos together with the corresponding CoT were sent as a prefix to the actual question. It is also worth to note that the examples were checked to be different from the test video in each iteration.

\paragraph{Text-based ablation.} LMMs strongly rely on the text context to answer questions. To establish a lower-bound for the LMMs' expected performance, we conducted text-based ablation on blind model performance. We tried removing completely the images from the input prompt, as well as replacing the video images with blank images. However, in both cases the models detected the missing visual input and did not provide a prediction. The models' response was something like:
"Actual images are not provided, therefore, this is a hypothetical analysis... Prediction: None".

\clearpage

\section{Detailed results}
\label{sec:supp_results_detail}

\paragraph{Event bounding-box detection.} We first examine a basic property: whether the model-predicted bounding box contains the true event location point.
\stabref{supptab:point-containment} reports, for each model, the fraction of ground-truth event points that fall inside the predicted bounding box.

\begin{table}[h]
\centering
\begin{tabular}{lcc}
\toprule
Model & Total label points in bbox & Percentage \\
\midrule
GPT-4o  & 56 / 99 & 56.57\% \\
GPT-5.2  & 86 / 99 & 86.87\% \\
Gemini-2.5-Flash  & 25 / 99  & 25.25\% \\
Gemini-3-Pro & 92 / 99 & 92.93 \% \\
Qwen-2.5VL & 99 / 99 & 100\%   \\
\bottomrule
\end{tabular}
\caption{Ground-truth event point locations contained inside the model-predicted bounding boxes.}
\label{supptab:point-containment}
\end{table}

\begin{figure*}[t]
    \centering
    \includegraphics[width=1\textwidth]{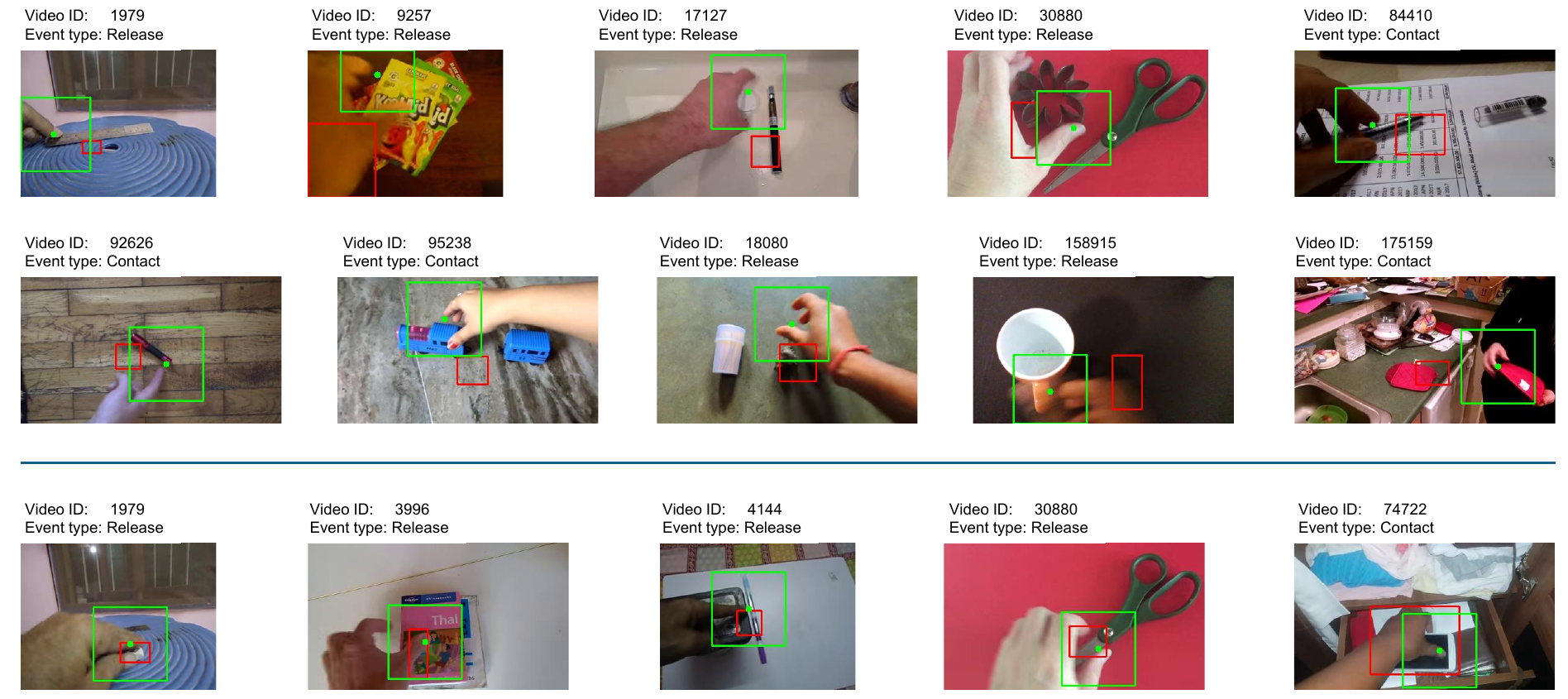}
    \caption{\textbf{Bounding box event detection examples of Gemini-3-Pro model.} Upper two rows: Failure detection examples. Bottom row: Successful detection examples. Green dot and bbox indicate the ground truth event location and a $120\times120$ pixels bboux around it. Red bbox indicates the model's prediction.}
    \label{sfig:supp_bbox_pred_examples}
\end{figure*}

Although the true event point often lies inside the predicted box (\stabref{supptab:point-containment}), the Intersection-over-Union (IoU) between the predicted box and a $120 \times 120$ pixels box around the true location point are extremely low (\stabref{supptab:iou-breakdown}, \sfigref{sfig:supp_bbox_pred_examples}). This discrepancy indicates that while the models often detect the hand or the target object, they tend to ignore important image regions around the target objects and hands, which contain critical information about the interactions. In some cases the predicted bounding box extended beyond the frame boundaries despite explicit instructions to preserve the original spatial scale (see \cref{supplst:bbox_prompt}). Overall, the models struggle to localize the interaction event regions reliably.

\begin{table}[h]
\centering
\begin{tabular}{lcc}
\toprule
Model & Mean IoU (\%) & \#IoU $\geq 50\%$ / 99 \\
\midrule
GPT-4o   & 1.48\% & 0 / 99 \\
GPT-5.2  & 8.62\%  & 0 / 99 \\
Gemini-2.5-Flash  & 0.21\%  & 0 / 99 \\
Gemini-3-Pro & 14.74\%  & 2 / 99 \\
Qwen-2.5VL  & 9.55\%  & 1 / 99 \\

\bottomrule
\end{tabular}
\caption{Per-model IoU statistics for the event localization task.}
\label{supptab:iou-breakdown}
\end{table}

\paragraph{\textit{Reasoning} and \textit{Grounding} conditions.} \stabref{stab:qwen_results_rsn} and \stabref{stab:qwen_results_grnd} complements the results for the models Qwen-2.5VL-72B and GPT-4o from our experiments on the two modifying conditions: \textit{Reasoning} and \textit{Grounding}.

\setlength{\tabcolsep}{3pt}
\begin{table}[t]
    \footnotesize
    \caption{\textbf{Reasoning effect on LMMs' performance.} Models are instructed to describe step-by-step the reasoning behind their prediction of the frame where the interaction occurs. Mean accuracy is measured for the detection of the frame where the test event occurs, within an allowed error tolerance (here the exact or one-frame off the true frame). Results are reported as per 3 ICL regimes (ZS, OS, TS) for the models: (i) Qwen-2.5VL-72B and (ii) GPT-4o.}
    \centering
            \begin{tabularx}{1.0\linewidth}{l|c|cc|cc}
            \toprule
                \textbf{ICL}                 & \textbf{RSN} & \multicolumn{4}{c}{\textbf{Mean Accuracy Percentage (SD)}} \\ 
                & & \multicolumn{2}{c|}{\textbf{Exact}} & \multicolumn{2}{c}{\textbf{1-off}} \\
                & & \multicolumn{1}{c}{\textbf{Qwen}} & \multicolumn{1}{c|}{\textbf{GPT-4o}} & \multicolumn{1}{c}{\textbf{Qwen}} & \multicolumn{1}{c}{\textbf{GPT-4o}} \\
                \midrule
                \multirow{2}*{\textbf{ZS}}  & W/O  & $7.07$ $(1.83)$ & $9.60$ $(2.10)$ & $31.31$ $(3.30)$ & $28.79$ $(3.23)$ \\
                & W  & $\textbf{10.61}$ $(2.19)$ & $9.09$ $(2.10)$  & $28.79$ $(3.23)$  & $\textbf{31.82}$ $(3.32)$ \\
                \midrule
                \multirow{2}*{\textbf{OS}}    & W/O   & $6.21$ $(1.38)$ & $\textbf{12.34}$ $(1.55)$ & $29.81$ $(2.81)$ & $\textbf{36.97}$ $(2.63)$ \\
                & W  &$11.53$ $(1.66)$ & $11.29$ $(1.20)$  & $36.12$ $(2.58)$ & $36.62$ $(2.36)$ \\
                \midrule
                \multirow{2}*{\textbf{TS}} & W/O & $10.39$ $(1.72)$ & $\textbf{14.72}$ $(1.63)$  & $36.44$ $(3.05)$ & $38.61$ $(2.86)$ \\
                & W   & $14.34$ $(1.90)$ & $13.44$ $(1.29)$ & $39.31$ $(2.76)$ & $\textbf{40.52}$ $(2.44)$ \\
                \bottomrule
            \end{tabularx}
    \label{stab:qwen_results_rsn}
\end{table}

\setlength{\tabcolsep}{3pt}
\begin{table}[t]
    \footnotesize
    \caption{\textbf{Grounding effect on LMMs' performance.} Prior to the main interaction detection task, models are instructed to name the target object and specify the length of the video sequence, to improve their perceptual grounding. Mean accuracy is measured for the detection of the frame where the test event occurs, within an allowed error tolerance (here the exact true frame or up to one- frame off). Results are reported as per 3 ICL regimes (ZS, OS, TS) for the models: (i) Qwen-2.5VL-72B and (ii) GPT-4o.}
    \centering
            \begin{tabularx}{1.0\linewidth}{l|c|cc|cc}
            \toprule
                \textbf{ICL}                 & \textbf{GRND} & \multicolumn{4}{c}{\textbf{Mean Accuracy Percentage (SD)}} \\
                & & \multicolumn{2}{c|}{\textbf{Exact}} & \multicolumn{2}{c}{\textbf{1-off}} \\
                & & \multicolumn{1}{c}{\textbf{Qwen}} & \multicolumn{1}{c|}{\textbf{GPT-4o}} & \multicolumn{1}{c}{\textbf{Qwen}} & \multicolumn{1}{c}{\textbf{GPT-4o}} \\
                \midrule
                \multirow{2}*{\textbf{ZS}}   & W/O  & $9.60$ $(2.10)$ & $\textbf{10.61}$ $(2.19)$ & $30.30$ $(3.27)$ & $\textbf{32.32}$ $(3.33)$  \\
                & W  & $8.08$ $(1.94)$ & $8.08$ $(1.94)$  & $29.80$ $(3.26)$  & $28.28$ $(3.21)$ \\
                \midrule
                \multirow{2}*{\textbf{OS}}    & W/O & $9.40$ $(1.58)$ & $\textbf{12.13}$ $(1.39)$  & $34.59$ $(2.77)$ & $\textbf{37.29}$ $(1.39)$ \\
                & W  & $8.34$ $(1.50)$ & $11.49$ $(1.39)$   & $31.34$ $(2.63)$ & $36.49$ $(3.31)$ \\
                \midrule
                \multirow{2}*{\textbf{TS}} & W/O & $13.11$ $(1.72)$ & $14.04$ $(1.44)$ & $38.92$ $(2.94)$ & $\textbf{42.06}$ $(2.57)$ \\
                & W & $11.63$ $(1.77)$ & $\textbf{14.11}$ $(1.49)$  & $36.83$ $(2.89)$ & $37.07$ $(2.73)$ \\
                \bottomrule
            \end{tabularx}
    \label{stab:qwen_results_grnd}
\end{table}

\begin{figure*}[t]
    \centering
    \begin{subfigure}{0.9\linewidth}
        \centering
        \includegraphics[width=1\textwidth]{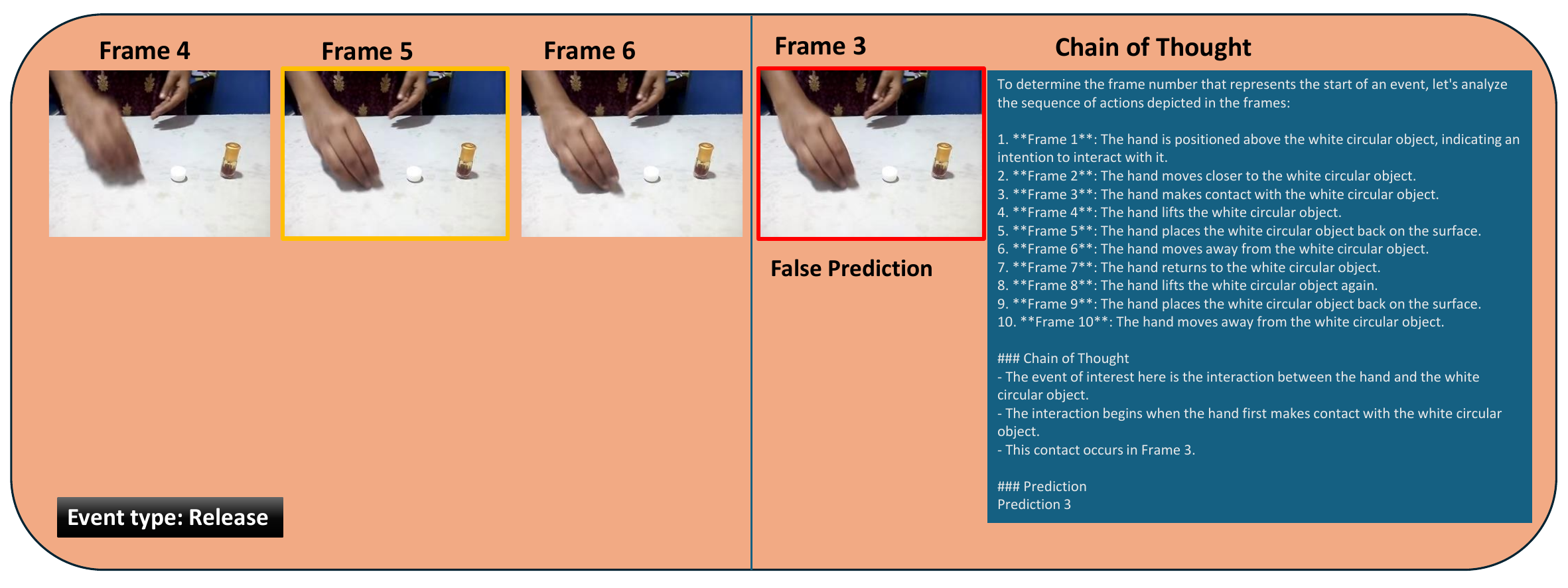}
    \end{subfigure}%
    \par 
    \begin{subfigure}{0.9\linewidth}
        \centering
        \includegraphics[width=1\textwidth]{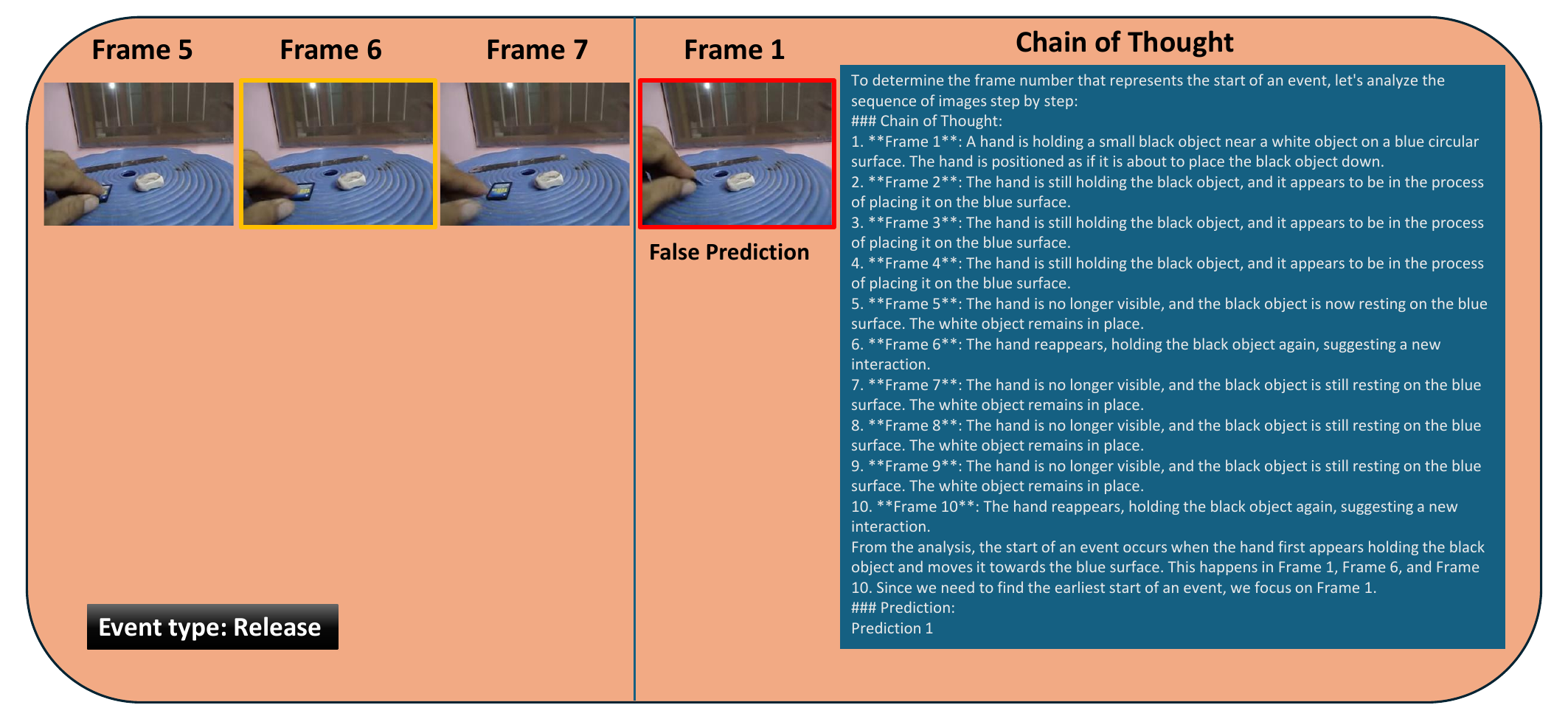}
    \end{subfigure}
    \par 
    \begin{subfigure}{0.9\linewidth}
        \centering
        \includegraphics[width=1\textwidth]{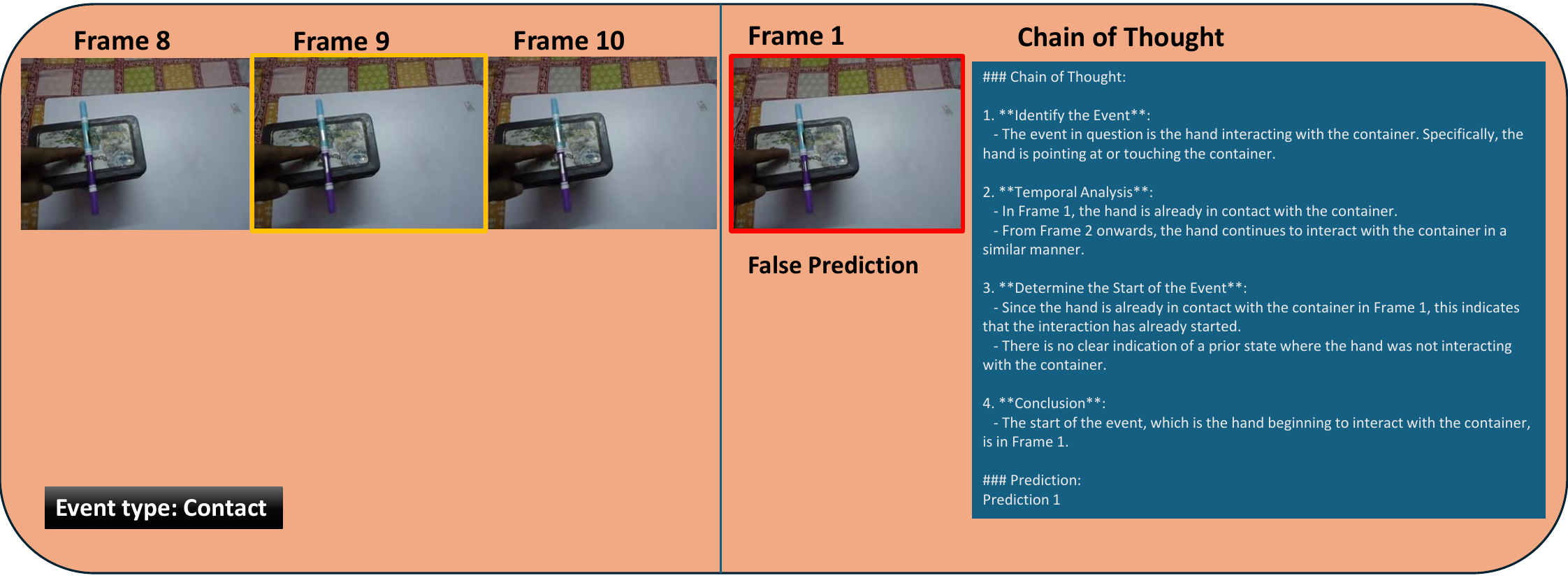}
    \end{subfigure}
    \caption{\textbf{Example false predictions.} The model (here Qwen-2.5VL-72B) provides the presented chain-of-thought under the "WITH" \textit{Reasoning} condition. The examples show that the reasoning text seems logical and realistic, but the relation to the actual video frames is often very loose. A red box marks a false prediction, while the orange box marks the true frame.}
    \label{sfig:supp_pred_false}
\end{figure*}

\begin{figure*}[t]
    \centering
    \begin{subfigure}{\linewidth}
        \centering
        \includegraphics[width=1\textwidth]{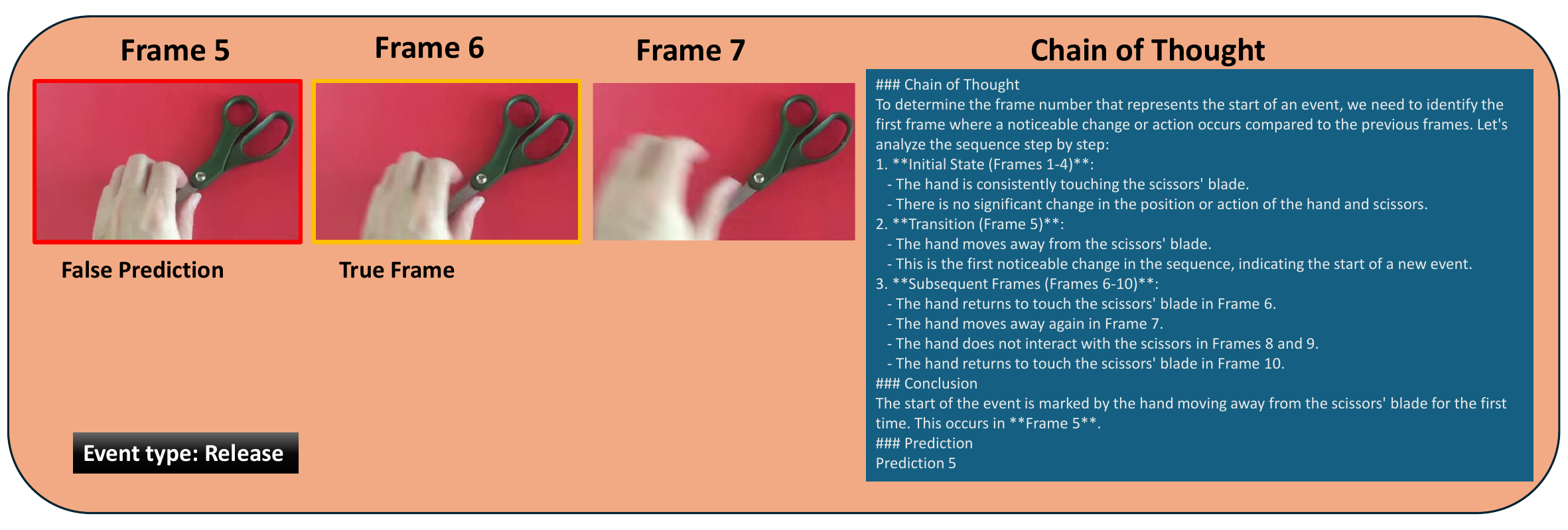}
    \end{subfigure}%
    \par 
    \begin{subfigure}{\linewidth}
        \centering
        \includegraphics[width=1\textwidth]{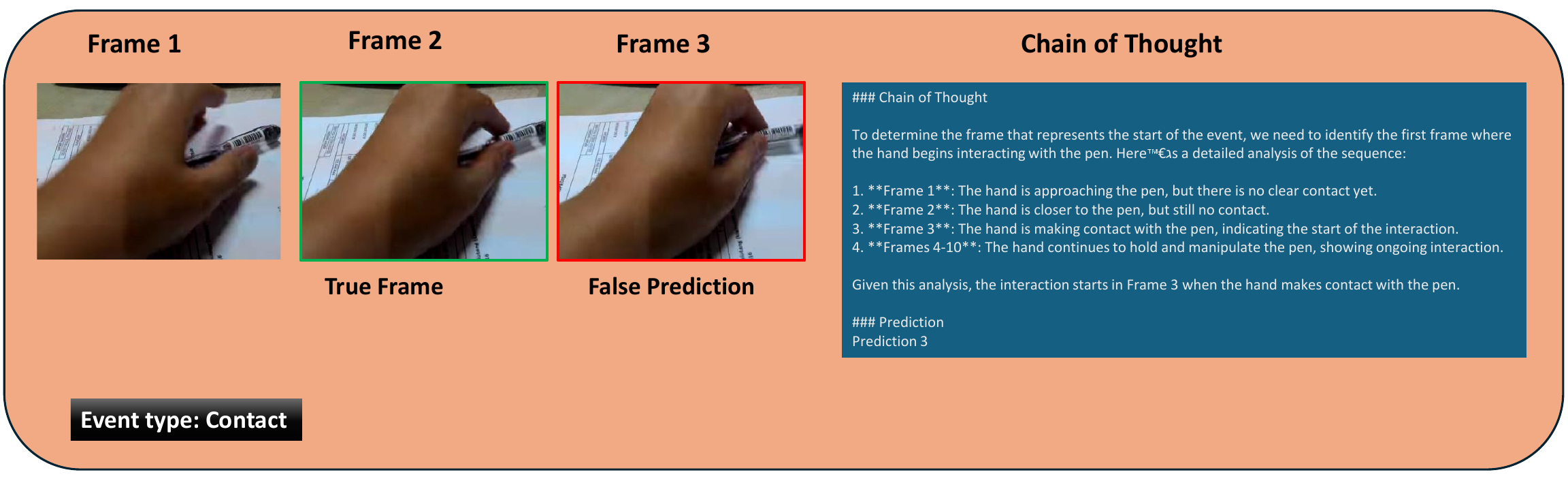}
    \end{subfigure}
    \par 
    \begin{subfigure}{\linewidth}
        \centering
        \includegraphics[width=1\textwidth]{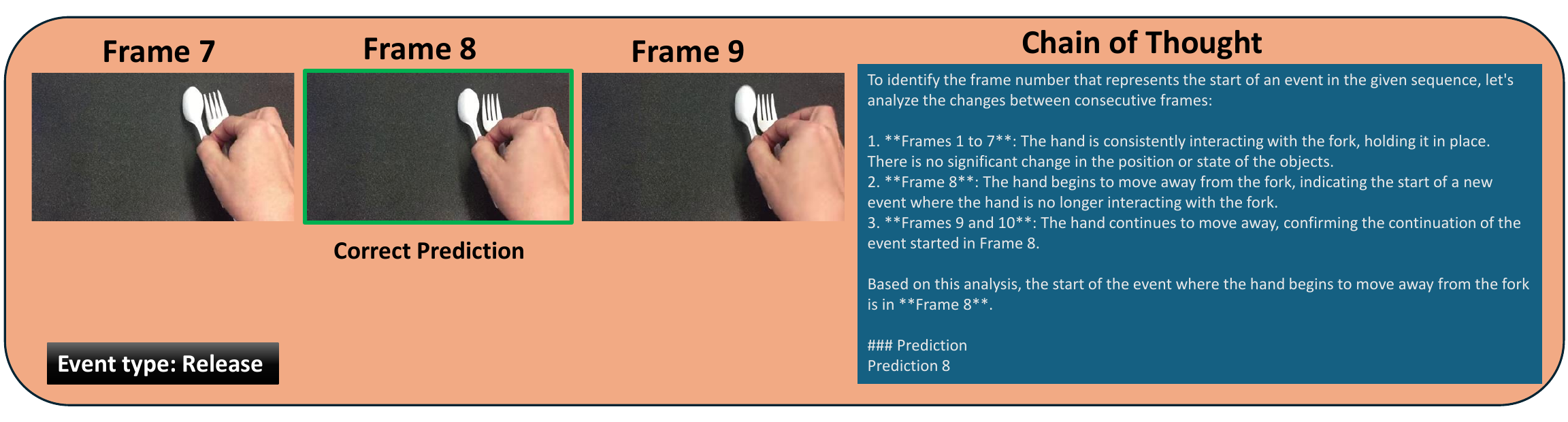}
    \end{subfigure}
    \caption{\textbf{Additional example predictions.} The top two examples show false predictions of the model (here Qwen-2.5VL-72B), in which the predicted frame was one frame before or after the true frame, where the event occurs. Humans can see clearly the moment of release or contact in the true frames, but the visual cues are too subtle for the models to detect. The Chain-of-Thought text lists the correct reasoning flow, but is not well grounded in the video frames. Orange and green boxes mark the true frame. Red boxes mark false predictions.}
    \label{sfig:supp_pred_true}
\end{figure*}

\paragraph{Example predictions and associated CoT.} \sfigref{sfig:supp_pred_false} and \sfigref{sfig:supp_pred_true} show additional examples of false and correct frame predictions. The examples include the models' Chain-of-Thought, which seemingly presents a logical reasoning text for detecting an interaction event, but the grounding to the video frames is often very loose.

\paragraph{Two-shot with feedback.} The results of this experiment showed no improvement in the model's performance on the test task of predicting the frame where an event occurs in the test video. The mean accuracy is reported in \cref{supptab:results_feedback}. The results suggest that the feedback session may even interfere with the main prediction task, by shifting away the model from the visual input to the number of the frame, while trying to maximize the feedback score, which is a metric on the prediction error of the frame number.

\setlength{\tabcolsep}{6pt}
\begin{table}[t]
    \caption{\textbf{Effect of feedback on the model's performance.} The true label of the second example is provided through an iterative session with feedback, indicating the gap between the predicted and the true frame. Mean accuracy is measured for the detection of the frame where the test event occurs, within an allowed error tolerance (here the exact true frame or up to one-frame off). The evaluation is performed under the two modifying conditions: \textit{Reasoning} and \textit{Grounding}. The evaluated model is Qwen-2.5VL-72B.}
    \centering
        \begin{tabularx}{1.0\linewidth}{c|c|c|c}
        \toprule
            \textbf{Reasoning} & \textbf{Grounding} & \multicolumn{2}{c}{\textbf{Mean Accuracy (\%)}} \\
                & &\multicolumn{1}{c|}{\textbf{Exact}} & \multicolumn{1}{c}{\textbf{1-off}} \\ 
                \midrule
                W/O & W/O &$2.55\pm1.12$ & $18.37\pm3.43$       \\
                W/O & W & $5.61\pm1.76$ & $21.94\pm3.70$         \\ 
                W & W/O & $12.59\pm2.63$ & $\textbf{41.84}\pm4.15$                \\ 
                W & W &  $\textbf{16.33}\pm3.06$ & $39.80\pm4.10$               \\ 
                \bottomrule
            \end{tabularx}
    \label{supptab:results_feedback}
\end{table}

\begin{figure*}[h]
\begin{subfigure}[b]{0.48\linewidth}
    \centering
    \includegraphics[width=1\linewidth]{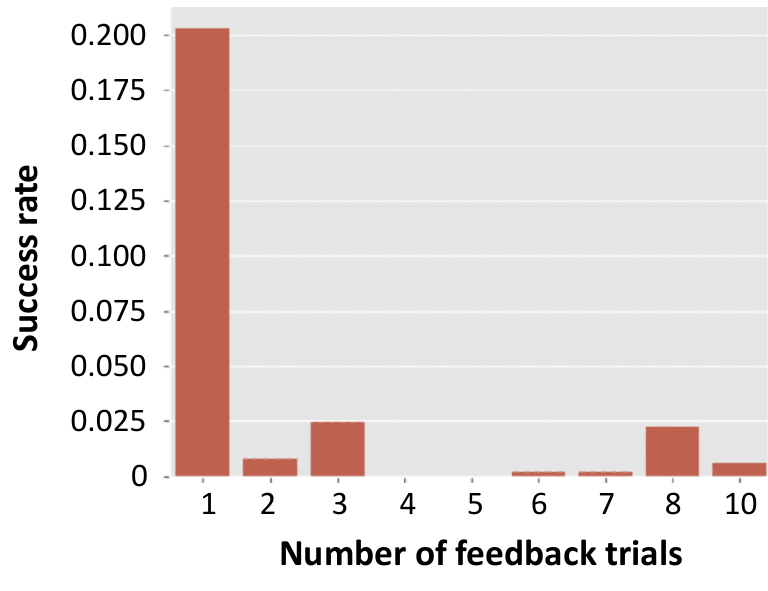} 
    \caption{Test prediction success rate - Without \textit{Reasoning}} 
    \vspace{4ex}
   \end{subfigure}
  \begin{subfigure}[b]{0.48\linewidth}
    \centering
    \includegraphics[width=1\linewidth]{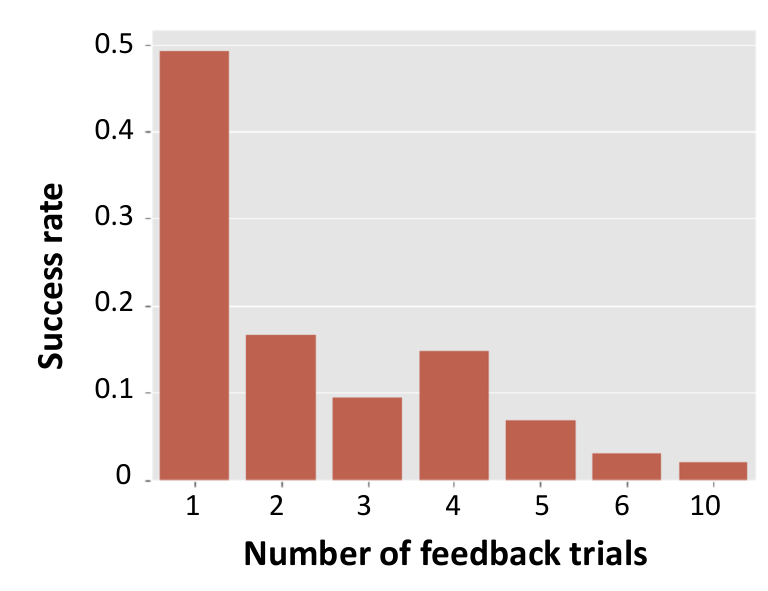} 
    \caption{Test prediction success rate - With \textit{Reasoning}} 
    \vspace{4ex}
  \end{subfigure}
  \begin{subfigure}[b]{0.48\linewidth}
    \centering
    \includegraphics[width=1\linewidth]{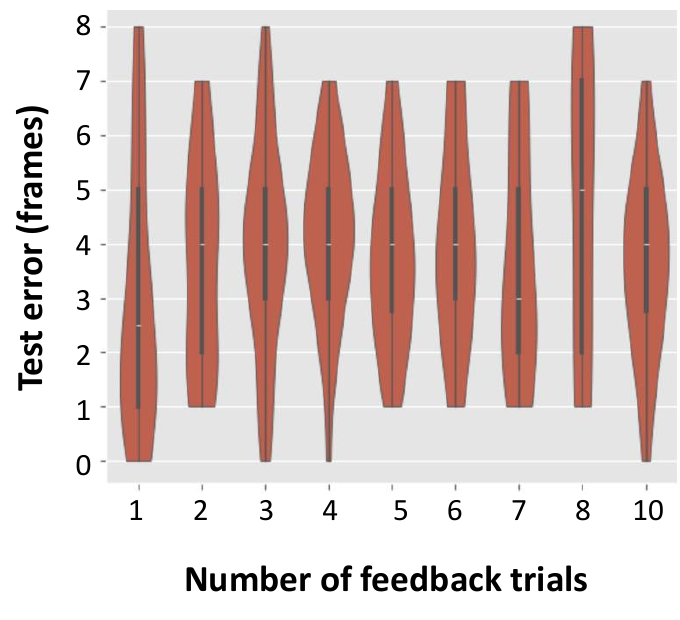} 
    \caption{Distribution of test error (frames) - Without \textit{Reasoning}} 
    \vspace{4ex}
   \end{subfigure}
  \begin{subfigure}[b]{0.48\linewidth}
    \centering
    \includegraphics[width=1\linewidth]{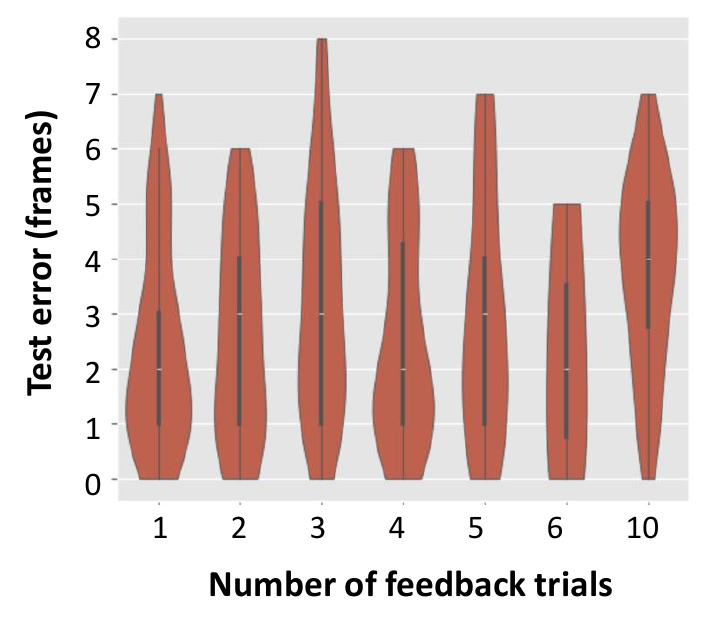} 
    \caption{Distribution of test error (frames) - With \textit{Reasoning}}
    \vspace{4ex}
   \end{subfigure}
   \caption{\textbf{Effect of feedback on test success rate and error distribution.} In the TS with feedback experiment, the model goes through an iterative session, when it is provided with a feedback score until it predicts correctly the event frame of the second example. (a-b) The model's weighted success rate (see \cref{sec:supp_results_detail}) on the test video vs. the number of trials in the feedback session, without and with \textit{Reasoning}, respectively. (c-d) The distribution of the test error (in frames) without and with \textit{Reasoning}, respectively. The results show that when the model is not required to provide reasoning for its prediction, the feedback session does not improve the model's performance in the test task and the test error increases with the number of feedback trials (a, c). On the other hand, when the model is required to provide reasoning, the test success rate decreases monotonically with the number of feedback trials up till the 6th iteration.}
\label{sfig:supp_feedback_results}
\end{figure*}

Nevertheless, an analysis of the model's weighted success rate and its test error shown in \cref{sfig:supp_feedback_results}, indicate that when the model is required to provide reasoning which lead to its predictions, the test error remains low up until the 6th feedback iterations, suggesting that some learning may occur. However, as discussed in the main text, the loose grounding to the visual input, becomes even worst with this feedback approach since the focus of the model is drawn away from the image contents, trying to satisfy the feedback score around the frame number, rather than grounded visual cues.

The weighted success rate of a test trial (see \cref{sfig:supp_feedback_results}a,b) was calculated with the conditioned probability as follows
\begin{equation}
    p(success = n | iterations = k) = \frac{n}{N} \frac{n}{N_k} = \frac{n^2}{NN_k}
\end{equation}
\noindent where N is the total number of successful test trials (i.e., where the agent predicted the correct frame in the test task), and $N_k$ is the number of successful test predictions conditioned on training session having $k$ iterations.   

It should be noted that we also tested a setting in where the images were fed to the model in an arbitrary order. This change in image ordering alone resulted in a significant drop in performance, as seen in \stabref{supptab:results_feedback}, despite an explicit instruction that was given to the model that the frames should be treated in consecutive order. From this experiment we conclude that in the functioning of LMMs there is no notion of ordering of frames, unless they are fed in as an ordered sequence. 


\paragraph{Chain-of-Thought (CoT) Tuning.} In this ICL few-shot experiment, examples included specific CoT descriptions for detecting the frame where the contact/release events occurred. The results in \stabref{stab:cot_res} show that the accuracy does not improve beyond 2 examples. However, introducing an explicit CoT description in the examples yielded enhanced accuracy compared to the experiment without CoT (see \stabref{stab:qwen_results_rsn}).  

\begin{table}[t]
\caption{Results for the CoT experiment expressed in mean accuracy within an error tolerance of one-frame off from the true frame in each test sample. The evaluated model is Qwen-2.5VL-72B. }
\centering
\begin{tabular}{cc}
\toprule
Number of Examples & Mean Accuracy (\%)\\
\midrule
1   & $51.52 \pm 5.02$ \\
2   & $55.56 \pm 4.99$ \\
3   & $45.45 \pm 5.00$\\
4   & $44.44 \pm 4.99$ \\
5   & $39.39 \pm 4.91$ \\
6   & $43.43 \pm 4.98$ \\
7   & $38.38 \pm 4.89$ \\
8   & $48.48 \pm 5.02$ \\
\end{tabular}
\label{stab:cot_res}
\end{table}

\begin{figure}[t]
    \centering
    \begin{subfigure}{\linewidth}
        \centering
        \caption{Qwen-2.5VL-72B}
        \includegraphics[width=1\textwidth]{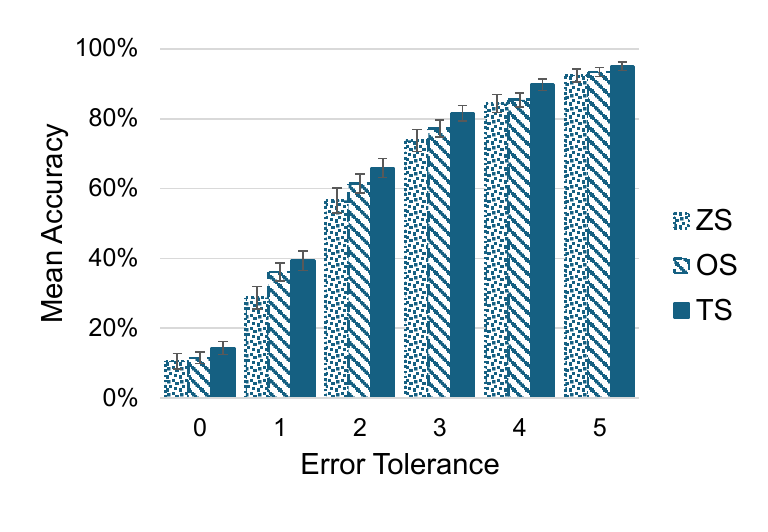}
    \end{subfigure}%
    \par 
    \begin{subfigure}{\linewidth}
        \centering
        \caption{GPT-4o}
        \includegraphics[width=1\textwidth]{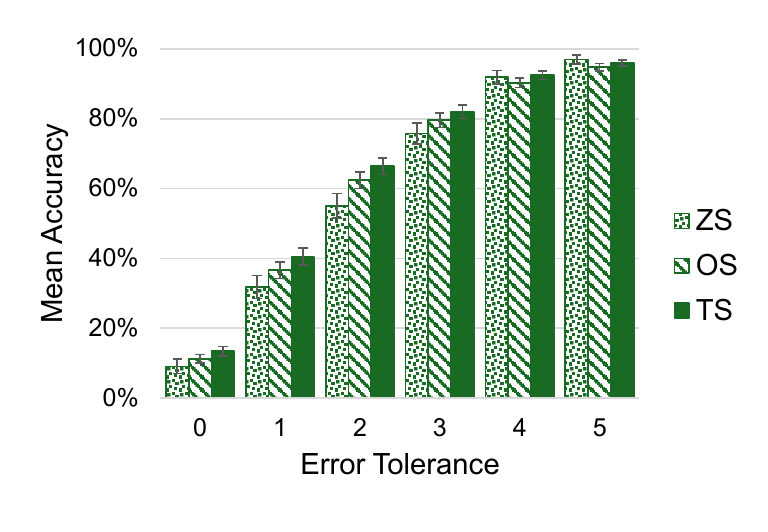}
    \end{subfigure}
    \caption{\textbf{Mean accuracy vs. detection error tolerance.} A correct detection of the models represents a predicted frame within the allowed error tolerance, where an error tolerance of zero means the exact true frame was predicted. Results of Qwen-2.5VL-72B (a) and GPT-4o (b) are shown for the difference ICL regimes under the "with reasoning" condition. Note, that the length of all videos in the experimental dataset is 10 frames.}
    \label{sfig:acc_vs_errtol}
\end{figure}

\section{Experimental Dataset}
\label{sec:supp_dataset_detail}
\begin{figure*}[t]
    \centering
    \includegraphics[width=1\textwidth]{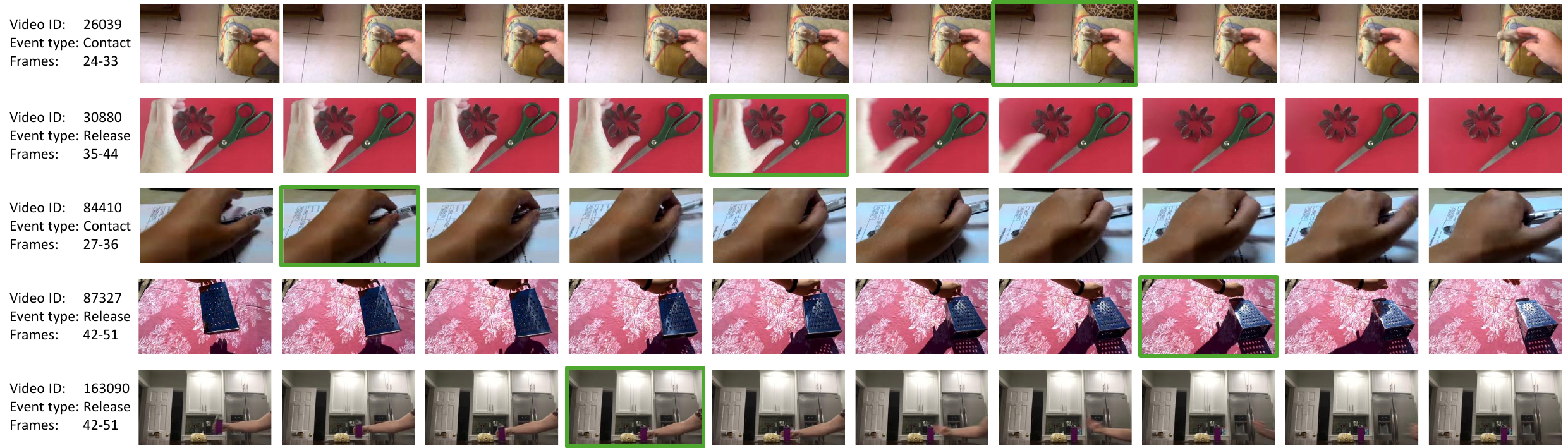}
    \caption{\textbf{Full interaction sequence examples from the evaluation dataset.} A green box marks the labeled event frame.}
    \label{sfig:supp_full_seq_examples}
\end{figure*}

Our experimental dataset included 33 videos from SSv2. For each video, we cropped short 10-frame video clips around three temporally separated core interaction events, resulting with 99 event clips. By construction, event frames are evenly distributed over the 10-frame window to avoid any bias. In addition, we have ensured that only a single event appears in the time window. \stabref{stab:exp_dataset_details} includes the video ID, action template and object placeholders from the original SSv2 dataset. In addition, for each short video clip, the table includes the crop start frame, the frame where the event occurs and the type of the event, i.e., 'contact' or 'release'. All video clips in this set are at 12 fps. \sfigref{sfig:supp_full_seq_examples} presents a few examples of the 10-frames clips and annotated event frames used in the evaluation.
The full annotations are available online at: \href{https://gitfront.io/r/hararid/52Roq1ASsNte/ssv2-contact-release-interaction-dataset}{ssv2-contact-release-interaction-dataset}.

\clearpage
\onecolumn

\setlength\LTleft{0pt}
\setlength\LTright{0pt}

{\footnotesize
\begin{longtable}{@{} c| L{5cm}| L{3cm}| c| c| c| c @{}}
    \caption{\textbf{Experimental dataset information.}}
    \label{stab:exp_dataset_details} \\
    \toprule
    \textbf{Video ID} & \textbf{Action template} & \textbf{Object placeholders} & \textbf{Frame size} 
    & \multicolumn{3}{c}{\textbf{Event clip\footnotemark} (10 frames)} \\
    & & & (width,height) & \multicolumn{1}{c|}{\textbf{Start frame}} 
          & \multicolumn{1}{c|}{\textbf{Event frame}} 
          & \multicolumn{1}{c}{\textbf{Event type}} \\
    \midrule
    \endfirsthead

    \caption[]{\textbf{.} (cont.)} \\
    \toprule
    \textbf{Video ID} & \textbf{Action template} & \textbf{Object placeholders} & \textbf{Frame size} 
    & \multicolumn{3}{c}{\textbf{Event clip} (10 frames)} \\
    & & & & \multicolumn{1}{c|}{\textbf{Start frame}} 
          & \multicolumn{1}{c|}{\textbf{Event frame}} 
          & \multicolumn{1}{c}{\textbf{Event type}} \\
    \midrule
    \endhead

    \midrule
    \multicolumn{6}{r}{\footnotesize\itshape Continued on the next page} \\
    \endfoot

    \bottomrule
    \endlastfoot

            \multirow{3}{*}{1979} & \multirow{3}{5cm}{Putting [something], [something] and [something] on the table} & \multirow{3}{3cm}{scale, eraser, sd card} & \multirow{3}{*}{$(320,240)$}
            & 12 & 14 & release \\
            & & & & 40 & 39 & release \\
            & & & & 60 & 55 & release \\
            \midrule
            \multirow{3}{*}{2648} & \multirow{3}{5cm}{Attaching [something] to [something]} & \multirow{3}{3cm}{dummy peach, peach tree} & \multirow{3}{*}{$(427,240)$}
            & 11 & 18 & contact \\
            & & & & 30 & 36 & contact \\
            & & & & 46 & 52 & release \\
            \midrule
            \multirow{3}{*}{3996} & \multirow{3}{5cm}{Putting number of [something] onto [something]} & \multirow{3}{3cm}{books, shelf} & \multirow{3}{*}{$(427,240)$}
            & 8 & 13 & release \\
            & & & & 29 & 33 & release \\
            & & & & 47 & 50 & release \\
            \midrule
            \multirow{3}{*}{4042} & \multirow{3}{5cm}{Pushing [something] so it spins} & \multirow{3}{3cm}{green candy}
            & \multirow{3}{*}{$(427,240)$} 
            & 7 & 10 & release \\
            & & & & 18 & 21 & contact \\
            & & & & 27 & 34 & release \\
            \midrule
            \multirow{3}{*}{4144} & \multirow{3}{5cm}{Poking [something] so that it falls over} & \multirow{3}{3cm}{pen}
            & \multirow{3}{*}{$(320,240)$}
            & 12 & 20 & contact \\
            & & & & 19 & 22 & release \\
            & & & & 32 & 34 & release \\
            \midrule
            \multirow{3}{*}{9257} & \multirow{3}{5cm}{Piling [something] up} & \multirow{3}{3cm}{kool-aid packs}
            & \multirow{3}{*}{$(320,240)$}
            & 8 & 14 & release \\
            & & & & 34 & 38 & release \\
            & & & & 52 & 57 & release \\
            \midrule
            \multirow{3}{*}{12492} & \multirow{3}{5cm}{Putting [something], [something] and [something] on the table} & \multirow{3}{3cm}{keys, lock, bulb} & \multirow{3}{*}{$(293,240)$}
            & 1 & 7 & release \\
            & & & & 10 & 17 & release \\
            & & & & 28 & 34 & release \\
            \midrule
            \multirow{3}{*}{14990} & \multirow{3}{5cm}{Putting [something], [something] and [something] on the table} & \multirow{3}{3cm}{perfume bottle, naphthalene ball, silver ring} & \multirow{3}{*}{$(320,240)$}
            & 22 & 25 & release \\
            & & & & 32 & 37 & release \\
            & & & & 48 & 52 & release \\
            \midrule
            \multirow{3}{*}{17127} & \multirow{3}{5cm}{Putting [something], [something] and [something] on the table} & \multirow{3}{3cm}{prescribers guide book, medicine bottle, vape pen} & \multirow{3}{*}{$(432,240)$}
            & 13 & 17 & release \\
            & & & & 36 & 39 & release \\
            & & & & 56 & 62 & release \\
            \midrule
            \multirow{3}{*}{26039} & \multirow{3}{5cm}{Pushing [something] so that it falls off the table} & \multirow{3}{3cm}{toy} & \multirow{3}{*}{$(427,240)$}
            & 1 & 9 & contact \\
            & & & & 13 & 20 & contact \\
            & & & & 24 & 30 & contact \\
            \midrule
            \multirow{3}{*}{30880} & \multirow{3}{5cm}{Putting [something], [something] and [something] on the table} & \multirow{3}{3cm}{scissors, cookie cutter, grater} & \multirow{3}{*}{$(427,240)$}
            & 12 & 17 & release \\
            & & & & 35 & 39 & release \\
            & & & & 56 & 62 & release \\
            \midrule
            \multirow{3}{*}{41434} & \multirow{3}{5cm}{Stacking [number of] [something]} & \multirow{3}{3cm}{3, coins} 
            & \multirow{3}{*}{$(293,240)$}
            & 0 & 4 & contact \\
            & & & & 17 & 19 & contact \\
            & & & & 37 & 43 & contact \\
            \midrule
            \multirow{3}{*}{57029} & \multirow{3}{5cm}{Taking [something] out of [something]} & \multirow{3}{3cm}{tools, toolbox} & \multirow{3}{*}{$(427,240)$}
            & 0 & 1 & contact \\
            & & & & 14 & 17 & contact \\
            & & & & 40 & 45 & contact \\
            \midrule
           \multirow{3}{*}{66464} & \multirow{3}{5cm}{Moving [part] of [something]} & \multirow{3}{3cm}{tuner, electric guitar}
           & \multirow{3}{*}{$(320,240)$}
            & 12 & 14 & contact \\
            & & & & 26 & 32 & release \\
            & & & & 36 & 40 & contact \\
            \midrule
            \multirow{3}{*}{67618} & \multirow{3}{5cm}{Putting [something], [something] and [something] on the table} & \multirow{3}{3cm}{bottle, tube, purse} & \multirow{3}{*}{$(427,240)$}
            & 0 & 1 & release \\
            & & & & 13 & 21 & release \\
            & & & & 40 & 47 & release \\
            \midrule
            \multirow{3}{*}{73232} & \multirow{3}{5cm}{Taking [something] out of [something]} & \multirow{3}{3cm}{cd, book}
            & \multirow{3}{*}{$(427,240)$}
            & 4 & 8 & contact \\
            & & & & 11 & 18 & contact \\
            & & & & 20 & 23 & release \\
            \midrule
            \multirow{3}{*}{74722} & \multirow{3}{5cm}{Taking [something] out of [something]} & \multirow{3}{3cm}{phone, drawer}
            & \multirow{3}{*}{$(427,240)$}
            & 0 & 8 & contact \\
            & & & & 24 & 29 & contact \\
            & & & & 52 & 56 & release \\
            & & &  &  &  \\
            \midrule
            \multirow{3}{*}{84410} & \multirow{3}{5cm}{Attaching [something] to [something]} & \multirow{3}{3cm}{pen's cover, pen} & \multirow{3}{*}{$(427,240)$}
            & 6 & 10 & contact \\
            & & & & 18 & 25 & release \\
            & & & & 27 & 28 & contact \\
            \midrule
            \multirow{3}{*}{87327} & \multirow{3}{5cm}{Putting [something], [something] and [something] on the table} & \multirow{3}{3cm}{grater, whisk, corkscrew} & \multirow{3}{*}{$(427,240)$}
            & 4 & 11 & release \\
            & & & & 28 & 31 & release \\
            & & & & 42 & 50 & release \\
            \midrule
            \multirow{3}{*}{92626} & \multirow{3}{5cm}{Poking [something] so that it spins around} & \multirow{3}{3cm}{flashlight} & \multirow{3}{*}{$(427,240)$}
            & 2 & 7 & contact \\
            & & & & 29 & 31 & release \\
            & & & & 41 & 44 & contact \\    
            \midrule
            \multirow{3}{*}{95238} & \multirow{3}{5cm}{Attaching [something] to [something]} & \multirow{3}{3cm}{toy train engine, its coach} & \multirow{3}{*}{$(427,240)$}
            & 3 & 7 & contact \\
            & & & & 13 & 18 & contact \\
            & & & & 40 & 43 & contact \\
            \midrule
            \multirow{3}{*}{96903} & \multirow{3}{5cm}{Rolling [something] on a flat surface} & \multirow{3}{3cm}{perfume}
            & \multirow{3}{*}{$(320,240)$}
            & 2 & 7 & contact \\
            & & & & 15 & 18 & contact \\
            & & & & 25 & 29 & contact \\
            \midrule
            \multirow{3}{*}{153413} & \multirow{3}{5cm}{Putting [something], [something] and [something] on the table} & \multirow{3}{3cm}{fork, spoon, dish} & \multirow{3}{*}{$(427,240)$}
            & 10 & 17 & release \\
            & & & & 27 & 33 & release \\
            & & & & 54 & 61 & release \\
            \midrule
            \multirow{3}{*}{158080} & \multirow{3}{5cm}{Putting [something], [something] and [something] on the table} & \multirow{3}{3cm}{toothpick container, showpiece, padlock} & \multirow{3}{*}{$(427,240)$}
            & 2 & 6 & release \\
            & & & & 21 & 27 & release \\
            & & & & 44 & 47 & release \\
            \midrule
            \multirow{3}{*}{158915} & \multirow{3}{5cm}{Putting [something], [something] and [something] on the table} & \multirow{3}{3cm}{mug, spoon, gum} & \multirow{3}{*}{$(427,240)$}
            & 1 & 5 & release \\
            & & & & 20 & 25 & release \\
            & & & & 31 & 34 & contact \\
            \midrule
            \multirow{3}{*}{163090} & \multirow{3}{5cm}{Putting [something], [something] and [something] on the table} & \multirow{3}{3cm}{popcorn, vicks vaporub bottle, purple water bottle} & \multirow{3}{*}{$(427,240)$}
            & 5 & 10 & release \\
            & & & & 26 & 29 & release \\
            & & & & 42 & 45 & release \\
            \midrule
            \multirow{3}{*}{164784} & \multirow{3}{5cm}{Pushing [something] so that it almost falls off but doesn't} & \multirow{3}{3cm}{roll} & \multirow{3}{*}{$(320,240)$}
            & 3 & 6 & contact \\
            & & & & 30 & 35 & release \\
            & & & & 43 & 48 & contact \\
            \midrule
            \multirow{3}{*}{166894} & \multirow{3}{5cm}{Poking a stack of [something] without the stack collapsing} & \multirow{3}{3cm}{lincoln logs} & \multirow{3}{*}{$(427,240)$}
            & 15 & 19 & contact \\
            & & & & 38 & 41 & release \\
            & & & & 46 & 51 & contact \\
            \midrule
            \multirow{3}{*}{175159} & \multirow{3}{5cm}{Stacking [number of] [something]} & \multirow{3}{3cm}{5, hot pads}
            & \multirow{3}{*}{$(427,240)$}
            & 6 & 9 & contact \\
            & & & & 20 & 26 & contact \\
            & & & & 39 & 41 & contact \\
            \midrule
            \multirow{3}{*}{175167} & \multirow{3}{5cm}{Piling [something] up} & \multirow{3}{3cm}{water color containers}
            & \multirow{3}{*}{$(427,240)$}
            & 4 & 8 & contact \\
            & & & & 25 & 31 & release \\
            & & & & 41 & 44 & contact \\     
            \midrule
            \multirow{3}{*}{181367} & \multirow{3}{5cm}{Piling [something] up} & \multirow{3}{3cm}{shoes}
            & \multirow{3}{*}{$(320,240)$}
            & 0 & 2 & contact \\
            & & & & 19 & 26 & release \\
            & & & & 29 & 30 & contact \\
            \midrule
            \multirow{3}{*}{186500} & \multirow{3}{5cm}{Pulling [something] onto [something]} & \multirow{3}{3cm}{nail clipper, envelope} & \multirow{3}{*}{$(427,240)$}
            & 11 & 17 & contact \\
            & & & & 21 & 26 & release \\
            & & & & 29 & 31 & contact \\
            \midrule
            \multirow{3}{*}{217743} & \multirow{3}{5cm}{Putting [something], [something] and [something] on the table} & \multirow{3}{3cm}{glass vase, child's shoe, coffee mug} & \multirow{3}{*}{$(427,240)$}
            & 21 & 24 & release \\
            & & & & 41 & 46 & release \\
            & & & & 58 & 65 & release \\     
\end{longtable}
\footnotetext{All video clips in this set are at 12 fps.}
} 
\twocolumn

\end{document}